\documentclass[runningheads]{llncs}
\usepackage{graphicx}
\usepackage{amsmath,amssymb} 
\usepackage{color}
\usepackage[width=122mm,left=12mm,paperwidth=146mm,height=193mm,top=12mm,paperheight=217mm]{geometry}
\usepackage{bm} 
\begin{document}
\pagestyle{headings}
\mainmatter
\def\ECCV14SubNumber{641}  

\title{Pairwise Rotation Hashing for High-dimensional Features} 

\titlerunning{Pairwise Rotation Hashing for High-dimensional Features}

\authorrunning{Pairwise Rotation Hashing for High-dimensional Features}

\author{Kohta Ishikawa \and Ikuro Sato \and Mitsuru Ambai}
\institute{Denso IT Laboratory, Inc. \\
              \email{kishikawa@d-itlab.co.jp}}

\maketitle


\begin{abstract}
Binary Hashing is widely used for effective approximate nearest neighbors search. Even though various binary hashing methods have been proposed, very few methods are feasible for extremely high-dimensional features often used in visual tasks today. We propose a novel highly sparse linear hashing method based on pairwise rotations. The encoding cost of the proposed algorithm is $\mathrm{O}(n \log n)$ for n-dimensional features, whereas that of the existing state-of-the-art method is typically $\mathrm{O}(n^2)$. The proposed method is also remarkably faster in the learning phase. Along with the efficiency, the retrieval accuracy is comparable to or slightly outperforming the state-of-the-art.
Pairwise rotations used in our method are formulated from an analytical study of the trade-off relationship between quantization error and entropy of binary codes. Although these hashing criteria are widely used in previous researches, its analytical behavior is rarely studied. All building blocks of our algorithm are based on the analytical solution, and it thus provides a fairly simple and efficient procedure.
\end{abstract}

\section{Introduction} \label{sec:introduction}
Approximate nearest neighbors (ANN) search is widely used in retrieval \cite{Shakhnarovich2006,Sivic2003,Torralba2008,Aiger2013}, and the scale of its database has been increasing rapidly in recent times. Furthermore, to achieve more accurate retrieval results, high-dimensional features such as Fisher Vectors \cite{Perronnin2007,Perronnin2010} and VLAD \cite{Jegou2012} are being used in the computer vision community. To achieve feasible retrieval with such features, highly efficient ANN search methods is necessarily needed. 

Vector Quantization based methods are widely used and actively studied for ANN. For high-dimensions, Product Quantization \cite{Jegou2011} and its family are the state-of-the-art methods \cite{Ge2013}. It reduces high-dimensional vector space into direct product of small subspaces. Then a clustering is applied for each subspace to obtain representative vectors (quantizers). Although product quantization based methods are applicable to high-dimensional features, it is still not easy to obtain good quantizer in some cases, and a random rotation often needed before PQ is expensive in high-dimensions. And the floating-point distance calculation needed for retrieval is also expensive compared to binary-based methods \cite{Gong2013}.

Binary hashing is one of the most commonly used techniques for efficient retrieval \cite{Jain2008,Wang2010,Wang2010b,He2012}, recognition \cite{Torralba2008a}, and other problems\cite{Jegou2012,Chaudhry2010}. It is a series of methods that transforms real-valued feature vectors into binary-valued ones. Binary-valued vectors are highly favorable for large-scale or high-dimensional tasks because they provide high memory efficiency and fast Hamming distance calculation. There are a lot of methods proposed. Major approaches are categorized as Vector Quantization (VQ) based methods \cite{Wang2010a,Norouzi2013,He2013,Ge2013}, hyperplane based linear methods \cite{Gong2013b,Liu2012a,Gong2013}, and nonlinear hashing function methods \cite{Weiss2008,Joly2011,Weiss2012,Liu2012b,Fan2013}.

A typical nonlinear method is Spectral Hashing \cite{Weiss2008}, whose hashing functions are nonlinear eigenfunctions derived from a distribution of data. Some family of Locality Sensitive Hashing uses nonlinear hashing functions\cite{Pauleve2010}. Kernelized approaches have also been proposed \cite{Kulis2009a,Kulis2009b,Raginsky2009}. 
Spectral Hashing, ordinary uses a uniform distribution to deriving analytical solution, and its precision is empirically lower compared to ITQ\cite{Gong2013b} and other state-of-the-art methods for non-uniformly distributed data. To overcome this difficulty, a kernel based approaches are proposed\cite{Chaudhry2010}. But it is difficult to apply to high-dimensions.
Recently proposed method called Spherical Hashing \cite{Heo2012} is a example of non-kernelized nonlinear method. Since its hashing function is hyper-sphere based, it also needs euclidean distance calculation for hashing. The computational cost is getting large for high-dimensional data.

Recently, a bilinear hashing method, which is called BPBC, that is feasible in high-dimensions was proposed \cite{Gong2013}. To our knowledge, this is the first binary hashing method that can treat 10K dimensions or higher. However, this method folds feature vectors and bilinearly rotates them in the folded space. It is unable to treat all of the Special Orthogonal group (Rotational group) $SO(n)$. There is still no linear high dimensional binary hashing method that can directly treat $SO(n)$.

In this paper, we propose a new highly efficient linear binary hashing method. Our method is inspired by Isotropic Hashing \cite{Kong2012}. We found out its natural extension. First, we study the meaning of isotropic transformation analytically. Then we develop a efficient isotropic hashing algorithm and its extension using trade-off relationship between isotropy and entropy. 
Recently proposed Sparse Isotopic Hashing method\cite{Sato2013} produces sparse rotational matrices that yield isotropic variances;  however the learning of high-dimensional rotational matrices is not feasible in practice.
Our main contributions are, \newline \newline
\textbf{1: State-of-the-art computational cost and accuracy}

Our algorithm takes $\mathrm{O}(n \log n)$ encoding cost for $n$ dimensional features. The previously known state-of-the-art method BPBC requires $\mathrm{O}(n^2/d + nd)$ (typically $d=128$, with no dimension reduction case) cost. We show that the proposed algorithm is more accurate than BPBC. Moreover, it is remarkably faster in learning phase. The main cost consuming point of our algorithm is calculation of a variance-covariance matrix.  
We only need $\mathrm{O}(n^2 \log n)$ computational cost in learning iteration loop, whereas BPBC requires $\mathrm{O}(n^2 \log n)$ in each iteration step. Therefore it is practically faster than BPBC, although total computational cost of learning has the same order $\mathrm{O}(mn^2)$ in our algorithm and BPBC with $m$ training data size. \newline \newline
\textbf{2: Analytical treatment of hashing criteria}

Typical criteria for measuring hashing performance are quantization error, variance of each bit, and entropy. To the author's knowledge, analytical treatment of these criteria has not been well studied yet. We show an analytical result, and a set of algorithms derived naturally from the result. The analytical calculation is mainly based on gaussian distribution. That does not mean the proposed algorithm is only applicable to gaussian-distributed data. Because any non-gaussian distributions can be expanded around gaussian \cite{Kolassa2006}, assuming gaussian distribution means taking lowest order of expansion. As will be discussed below, lowest order approximation leads enough hashing accuracy and yields extremely efficient algorithm.

\section{Theoretical Background} \label{sec:theoretical_background}
\subsection{Quantization Error for Binary Hashing}
Along with clustering methods such as k-means clustering, most binary hashing algorithms aim at minimizing quantization error between binarized codes and original feature vectors\cite{Gong2013b,Gong2013}. In this study, therefore, the properties of quantization error are first investigated analytically. The result will be used to develop the binary hashing algorithm we propose.

Most of linear binary hashing methods consist of translation operation and linear transformation.
\begin{align} \label{eq:binhash}
&\bm{b}(\bm{x}) \equiv \text{sgn} (A (\bm{x} - \bm{t})), \\
\bm{x}, \bm{t} \in &\mathbb{R}^n, \bm{b}(\bm{x}) \in \{-1,1\}^m, A \in \mathbb{R}^{m \times n}. \nonumber
\end{align}
It is assumed that a translation is mean centering in the following discussion. 

Quantization error is defined as the sum of squared Euclidean distance between an original feature vector and its binarized vector.
\begin{align} \label{eq:e_q_emp}
E_{q} \equiv \frac{1}{N} \sum_i |\bm{x}_i - \bm{b}(\bm{x}_i)|^2,
\end{align}
where $N$ is the number of data points.
When the data is distributed as arbitrary distribution function $p(\bm{x})$, it is possible to write down mean quantization error. If two-dimensional data are assumed, 
\begin{align}
E_{q}^{N\to \infty} &= \int_0^{\infty} \!\!\! dx_1 \int_0^{\infty}\!\!\! dx_2 \left[(x_1 - 1)^2 + (x_2 - 1)^2\right] p(\bm{x}) \nonumber \\
&+ \int_{-\infty}^{0} \!\!\! dx_1 \int_0^{\infty}\!\!\! dx_2 \left[(x_1 + 1)^2 + (x_2 - 1)^2\right]p(\bm{x}) \nonumber \\
&+ \int_{-\infty}^{0}\!\!\! dx_1 \int_{-\infty}^{0}\!\!\! dx_2 \left[(x_1 + 1)^2 + (x_2 + 1)^2\right]p(\bm{x}) \nonumber \\
&+ \int_{0}^{\infty}\!\!\! dx_1 \int_{-\infty}^{0}\!\!\! dx_2 \left[(x_1 - 1)^2 + (x_2 + 1)^2\right]p(\bm{x}).
\end{align}
Then it is generally calculated as follows;
\begin{align}\label{eq:qerr_marginal}
&E_{q}^{N\to \infty} = \int_{-\infty}^{\infty}\!\!\! dx_1 \int_{-\infty}^{\infty}\!\!\! dx_2 \left[ x_1^2 + x_2^2 + 2\right] p(\bm{x}) \nonumber \\
&-2 \bigg[ \int_{0}^{\infty}\!\!\! dx_1 \, x_1 p_1(x_1) - \int_{-\infty}^{0}\!\!\! dx_1\, x_1 p_1(x_1) + \int_{0}^{\infty}\!\!\! dx_2 \, x_2 p_2(x_2) - \int_{-\infty}^{0}\!\!\! dx_2\, x_2 p_2(x_2) \bigg].
\end{align}
where $p_1(\cdot)$ and $p_2(\cdot)$ are marginal distributions with respect to $x_1$ and $x_2$.

If it is assumed that the distribution is gaussian with mean centering and variance-covariance matrix $\Sigma$, (\ref{eq:qerr_marginal}) is calculated as follows,
\begin{align} \label{eq:qerr_2d_gauss}
E_q^{N\to \infty} = 2 + \text{Tr}(\Sigma) - 2\sqrt{\frac{2}{\pi}} \left( \sqrt{\sigma_{11}} + \sqrt{\sigma_{22}} \right), \ \
\Sigma =
\begin{pmatrix}
\sigma_{11}& \sigma_{12} \\
\sigma_{12}& \sigma_{22}
\end{pmatrix},
\end{align}
where $\sqrt{\sigma_{11}}$ and $\sqrt{\sigma_{22}}$ are standard deviations of each dimension. These results are straightforwardly extended to higher dimensions case.

\subsection{Quantization Error Minimization in Gaussian Distribution with Rotational Transformation}
Some binary hashing algorithms use orthogonal transformation \cite{Gong2013b,Gong2013}. This means that their purpose is to find cost minimizing point in rotational group $SO(n)$. We also consider rotational group in this paper.

With rotational group transformation, minimizing (\ref{eq:qerr_2d_gauss}) is equivalent to maximizing $\sqrt{\sigma_{11}} + \sqrt{\sigma_{22}}$ subject to $\text{Tr}(\Sigma) = \text{const}$. Since constancy of $\text{Tr}(\Sigma)$ constrains a two dimensional vector $(\sqrt{\sigma_{11}},  \sqrt{\sigma_{22}})$ onto a circle, the solution is $\sigma_{11} = \sigma_{22}$. It is proved that the Isotropic Hashing \cite{Kong2012} is quantization-error-minimizing hashing for gaussian distribution. We can see the isotropy as a measure of quantization error.

\subsection{Entropy and Quantization Error} \label{sec:ent_qerr}

\begin{figure}[t]
\centering
\includegraphics[scale=0.23,angle=0,clip]{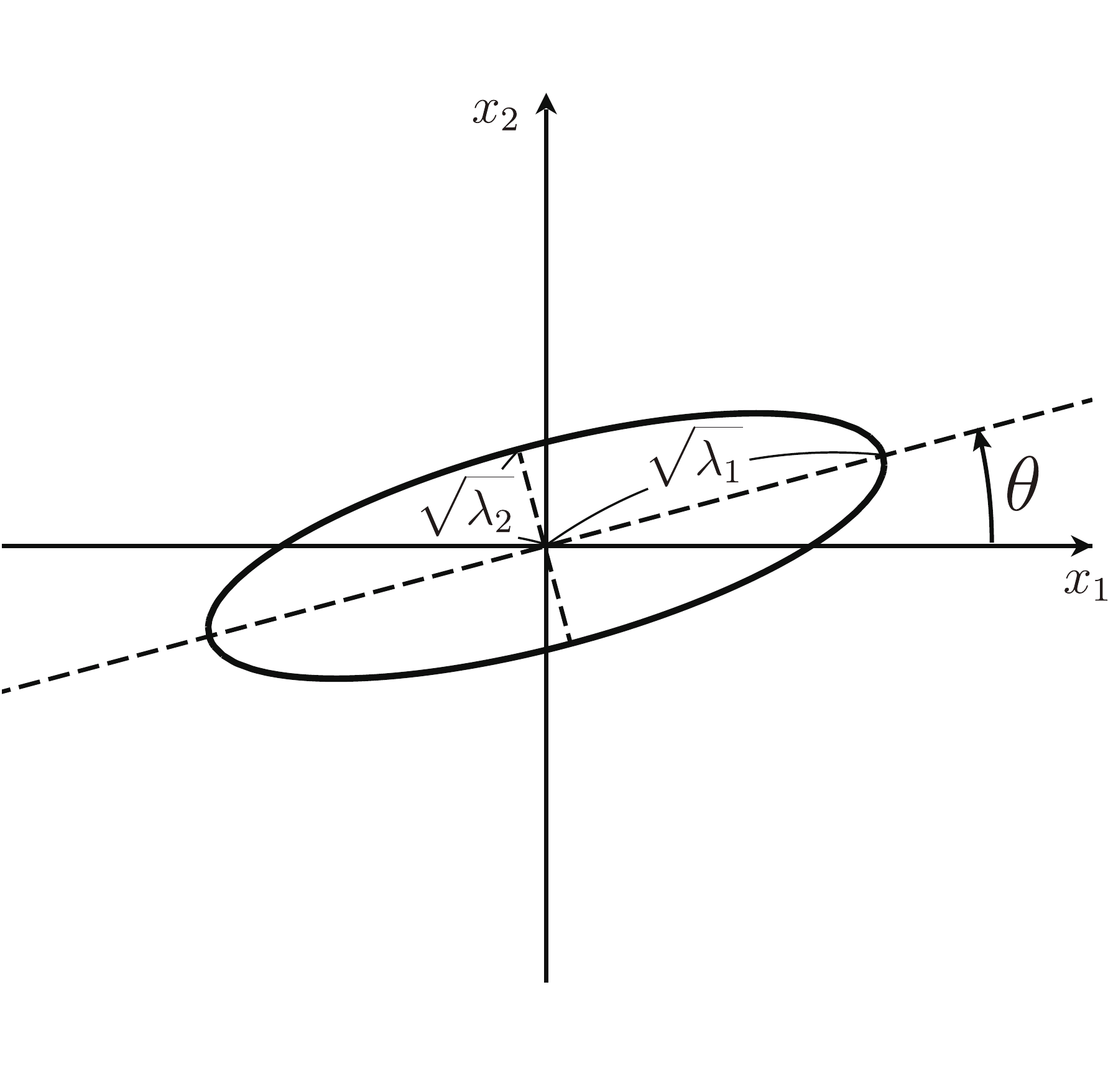}
\hspace{5mm}
\includegraphics[scale=0.23,angle=0,clip]{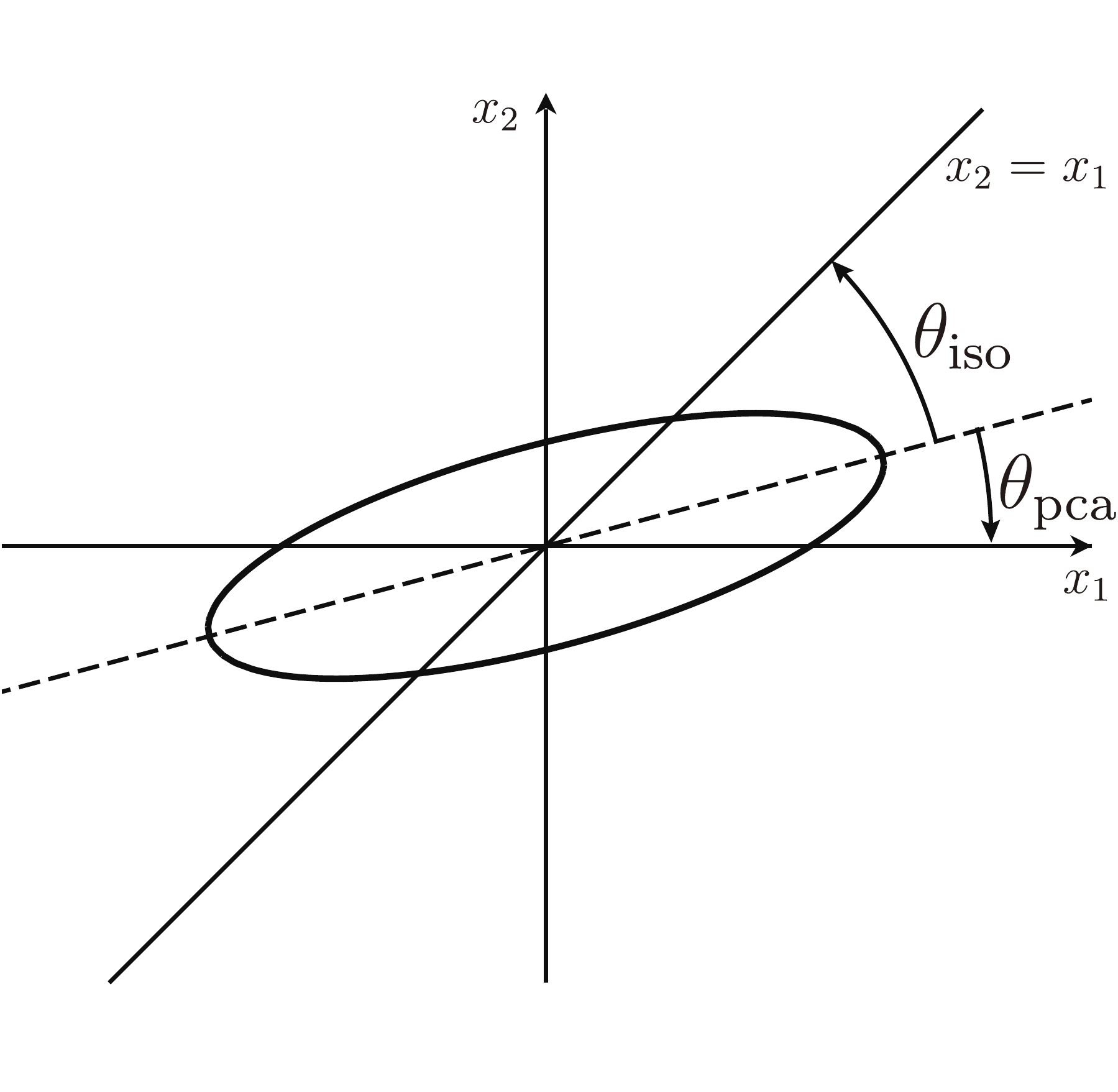}
\caption{Schematic illustration of two-dimensional gaussian distribution. An ellipse describes the shape of distribution. \textbf{Left:} eigenvalue and angle parameterization(Eq. (\ref{eq:r-theta_representation_of_2d_gauss})). $\lambda_1$ and $\lambda_2$ are the eigenvalues of variance-covariance matrix. \textbf{Right:} rotation angle for Isotropic or PCA transformation. $\theta_{\text{iso}}$ and $\theta_{\text{pca}}$ are given in (\ref{eq:iso_rotation}), (\ref{eq:pca_rotation}). \label{fig:ellipse}}
\end{figure}

\begin{figure}[t]
\centering
\includegraphics[scale=0.12,angle=0]{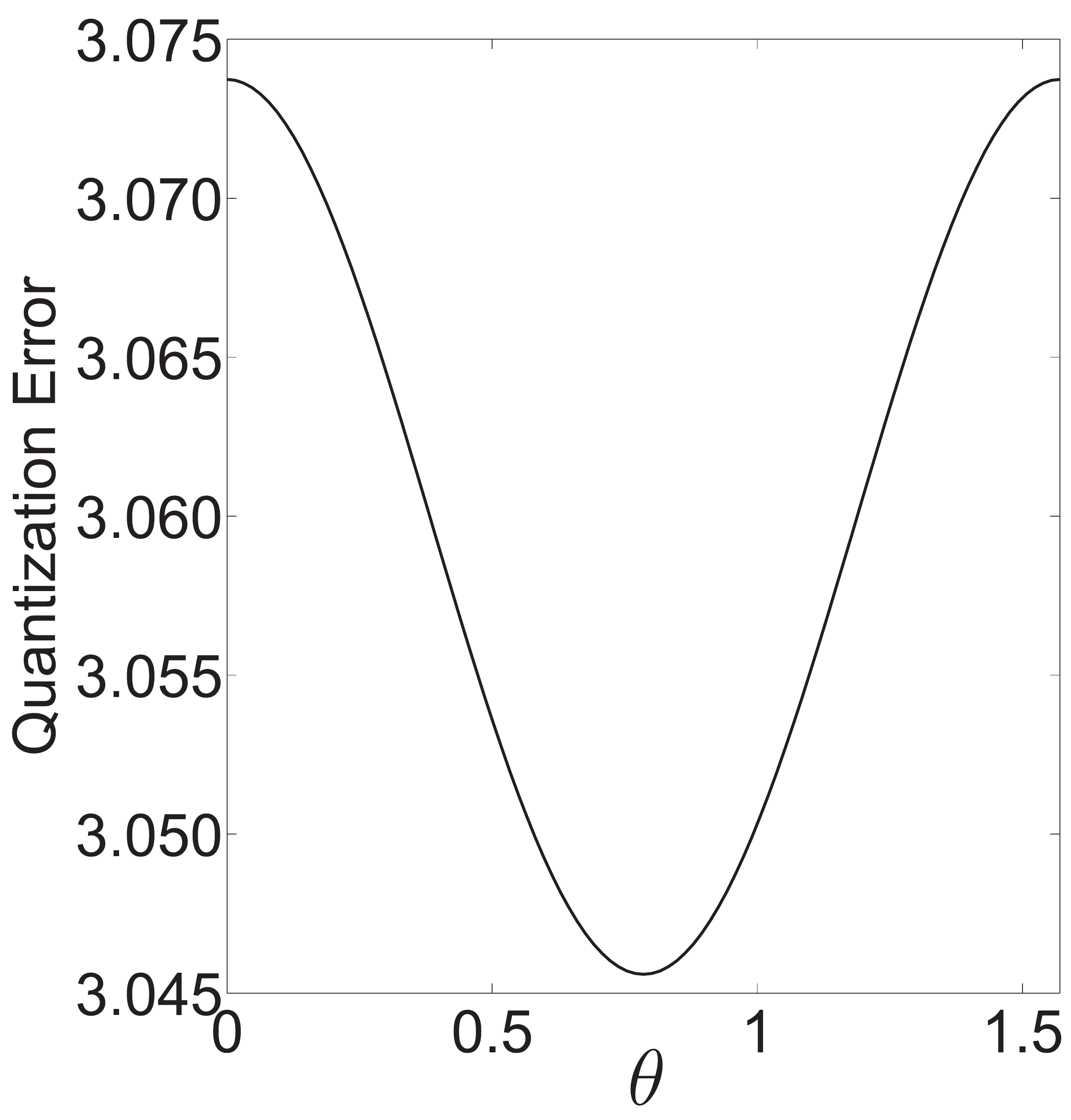}
\hspace{5mm}
\includegraphics[scale=0.12,angle=0]{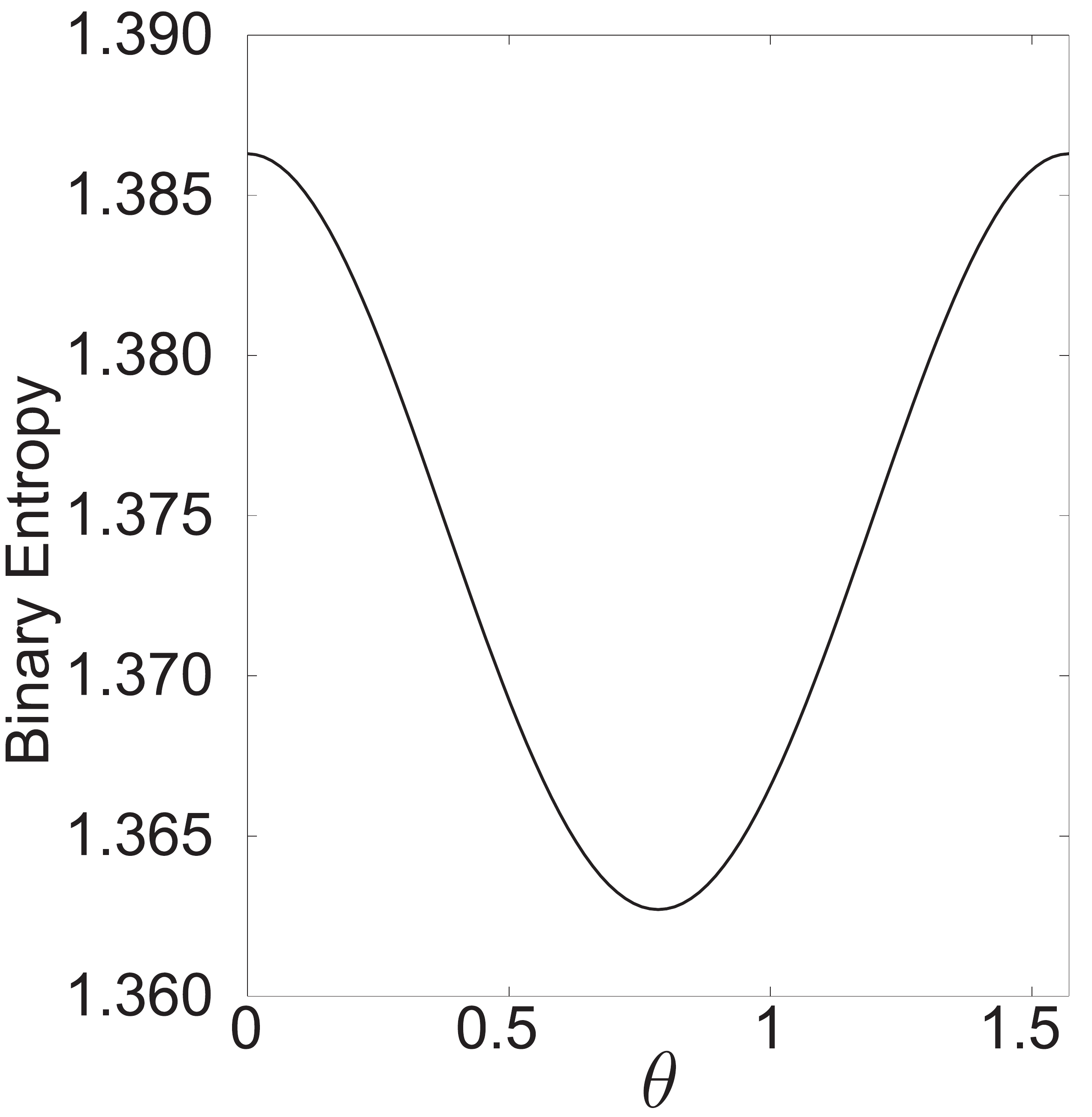}
\caption{Quantization error and entropy of binary code for $\lambda_1 = 2, \lambda_2 = 1$. The x-axis is angle $\theta$ given in equation (\ref{eq:r-theta_representation_of_2d_gauss}). From the symmetry of gaussian distribution, it is enough to consider range $\theta \in [0, \pi / 2]$ \label{fig:qerr_entropy}}
\end{figure}

Then the entropy of binary code is calculated. Here, eigenvalues and angle representation are used instead of a variance-covariance matrix, and only the two-dimensional case is treated. In this representation, elements of the variance-covariance matrix are described as follows;
\begin{align} \label{eq:r-theta_representation_of_2d_gauss}
\sigma_{11} = \bar{\lambda} + \lambda \cos &2\theta, \quad \sigma_{22} = \bar{\lambda} - \lambda \cos 2\theta, \quad 
\sigma_{12} = \lambda \sin 2\theta, \\
&\bar{\lambda} = \frac{\lambda_1 + \lambda_2}{2}, \quad \lambda = \frac{\lambda_1 - \lambda_2}{2}, \nonumber
\end{align}
where $\lambda_1$ and $\lambda_2$  are eigenvalues of variance-covariance matrix ($\lambda_1 \ge \lambda_2$). $\theta$ means the angle between $x_1$-axis and the longer axis direction of the gaussian ellipse (Fig. \ref{fig:ellipse}).

From the symmetry of gaussian distribution, it is enough to get probabilities of binary code $(1,1)$ and $(-1,1)$. It is possible to analytically calculate these probabilities as follows;
\begin{align}
&p_{(1,1)} = \int_0^{\infty} \!\!\! dx_1 \int_0^{\infty} \!\!\! dx_2 \ \frac{1}{2\pi} \frac{1}{|\Sigma|^{-1/2}} e^{-\frac{1}{2} \bm{x}^T \Sigma^{-1} \bm{x}}= \frac{1}{2} - \frac{1}{2\pi} \tan^{-1} \left( \frac{2}{\gamma \sin 2\theta}\right) \nonumber \\
&p_{(-1,1)} = \frac{1}{2} - p_{(1,1)} = \frac{1}{2\pi} \tan^{-1} \left( \frac{2}{\gamma \sin 2\theta}\right) \qquad \qquad \theta \in [0, \frac{\pi}{2}],
\end{align}
where $\gamma$ is defined as $\sqrt{\lambda_1 / \lambda_2} -\sqrt{ \lambda_2 / \lambda_1}$, which is the maximum value of correlation between $x_1$ and $x_2$ under rotational transformation.
The entropy of the two dimensional binary code is then given as
\begin{align}
S(\gamma, \theta) = 2 \times (- p_{(1,1)} \log p_{(1,1)} - p_{(-1,1)} \log p_{(-1,1)}).
\end{align}

Fig. \ref{fig:qerr_entropy} is plots of quantization error and entropy with respect to angle $\theta$. When quantization error is minimized, entropy is also minimized and vice versa. This means that the quantization error and the entropy have a trade-off relationship. Compatibility of the two factors depends on the "sharpness" of the distribution. When the distribution is sharp ($\lambda_1 \gg \lambda_2$), entropy is heavily damaged with isotropic variances. There is no trade-off relationship if the distribution is circular ($\lambda_1 = \lambda_2$). But in general case, we should balance these criteria.

An analysis that is similar to ours is recently proposed in \cite{Ge2013}. In the paper, the quantization error for Product Quantization was discussed. The authors showed that it is bounded by determinant of variance-covariance matrix, and proposed an algorithm that is minimizing the bound under rotational transformation. The result indicates trade-off relationship between quantization error and entropy of the gaussian distribution. However, since the entropy of the gaussian distribution is invariant under rotational transformation, this method only determines the partition of the entire space to the set of small subspaces. Rotations in each subspaces are not under discussion. By contrast, our analysis can consider rotational optimality in the two-dimensional subspaces because we investigate the entropy of binary codes directly.

Another example is \cite{Jin2013}. The authors proposed two criteria, that is the "crossing sparse region" and the "balanced buckets". The first criterion can be interpreted as quantization error minimization, and the second means quantizer entropy maximization. 
We think that it is possible to interpret many existing methods as such trade-off problem of quantization error minimization and entropy maximization.

\section{Methods}
A binary hashing algorithm based on the above-discussed theory is developed as follows. We are going to have very sparse transformation matrices, which substantially decrease encoding cost.
\subsection{Problem Statement}
The problem is to yield a linear transformation matrix $A$ in equation (\ref{eq:binhash}).  Most existing methods split the transformation into dimension reduction projection $W \in \mathbb{R}^{m \times n}$ and transformation in reduced space $Q \in \mathbb{R}^{m \times m}$.  PCA is commonly used for reducing the number of dimension. However, a PCA transformation matrix is dense, it is difficult to get transformation and efficient encoding calculation in highly dimensional cases. In this paper, dimension reduction is not treated. It is thus assumed that the number of dimensions of the original feature vector and the encoded binary vector are the same, and $A = Q \in \mathbb{R}^{n \times n}$ only is treated. Dimension reduction can be done in the similar way as we are going to discuss below, but detailed study is a future work.

\begin{figure}[t]
\centering
\includegraphics[scale=0.38,angle=0,clip]{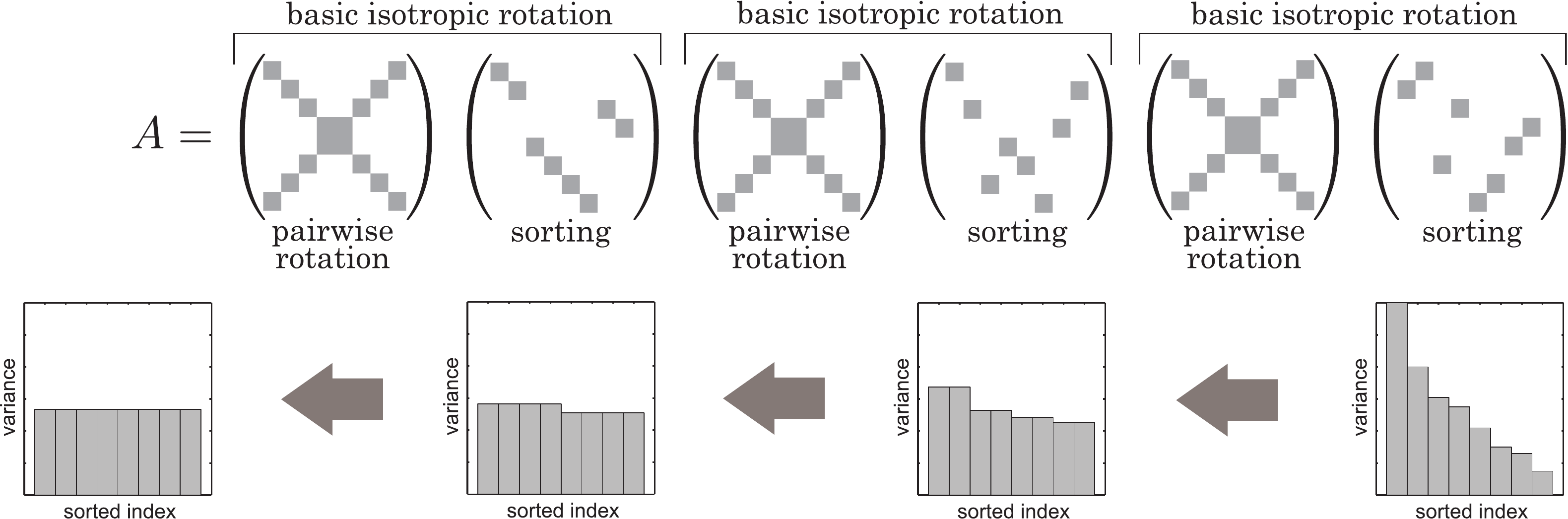}
\caption{Schematic illustration of basic isotropic rotations (in 8-dimensions). \textbf{Upper}: Structure of the transformation matrix. Non-zero matrix elements are filled with gray color. Sorting matrix is a permutation matrix, which sort variances in descending order. Rotation is done for pair of largest with smallest variance dimensions. Basic rotation is continuously applied $\log_2 n$ times for n dimensions.  \textbf{Lower}: Behavior of variances under continuous multiplication of basic isotropic rotation. The graphs are sorted variances under sequential multiplication. The rightmost graph is the (sorted) initial state. One basic isotropic rotation makes variances isotropic by pairs. And it is exponentially transformed to globally isotropic state with continuous application of basic rotation. \label{illustration:basic_rotation}}
\end{figure}

\subsection{Sequential Pairwise Isotropic Rotation} \label{sec:completely_isotropic}
First, we derive transformation that makes variances completely isotropic. We can get very sparse isotropic transformation matrix with $\mathrm{O}(n \log_2 n)$ fill-ins using pairwise isotropic rotation, although original Isotropic Hashing \cite{Kong2012} needs dense transformation matrix with $\mathrm{O}(n^2)$ fill-ins.

In two-dimensional space, there are only two isotropic transformations. It corresponds to $\theta = \pi/4, 3\pi/4$ in equation (\ref{eq:r-theta_representation_of_2d_gauss}). From the symmetry of the gaussian distribution, it is enough to consider $\theta = \pi/4$. For any two-dimensional variance-covariance matrix $\Sigma$, the rotation matrix that makes variances isotropic is
\begin{align} \label{eq:iso_rotation}
R = 
\begin{pmatrix}
\cos \theta_{\text{iso}} & -\sin \theta_{\text{iso}} \\
\sin \theta_{\text{iso}} & \cos \theta_{\text{iso}}
\end{pmatrix}
, \quad \theta_{\text{iso}} = \tan^{-1} \left( \frac{1}{2} \frac{\sigma_{11} - \sigma_{22}}{\sigma_{12}} \right).
\end{align}

To develop isotropic transformation for full dimension, we define "basic isotropic rotation". It consists of three steps. First step is to sort dimensions by diagonal elements of variance-covariance matrix in descending order. Second step is to create pairs of dimensions as $(1, n), (2, n-1), \cdots$. Third step is taking isotropic rotation (\ref{eq:iso_rotation}) for each pairs. The set of processes is denoted by a permutation (sorting) matrix and a rotational matrix with $2n$ fill-ins.  This transformation, which we call it "basic isotropic rotation"\footnotemark in what follows,  make variances pairwise isotropic.
\footnotetext{When a permutation is odd, the determinant of the transformation matrix is -1. In that case, the transformation is not a element of $SO(n)$, but $O(n)$. However, we can always obtain the element of $SO(n)$ by applying a odd permutation to the final matrix $A$. The application means a permutation of bits and does not affect retrieval results.}
Then we apply above transformation sequentially. Applying the transformation two times make variances quadruple isotropic, three times makes then octuple isotropic (Fig. \ref{illustration:basic_rotation}), and so on. Finally, applying the transformation $\lceil \log_2 n \rceil$ times, we have completely isotropic variances.
\footnote{In a precise sense, completely isotropic variances can be obtained for only $2^n$ dimensional case. For other dimensions, it needs infinite number of basic isotropic rotation. In practice, however, enough sub-isotropic variances can be obtained with $\lceil \log_2 n \rceil$ times transformation.} Finally developed transformation is a product of  sparse matrices. It has $2n\lceil \log_2 n \rceil$ fill-ins in total. This factorized form is highly sparse especially for high dimensions. It is possible to use standard sparse matrix data structure. Memory usage and computational cost can be substantially decreased.

\subsection{Trade-off between Quantization Error and Entropy}
A factorized sparse transformation that makes variances completely isotropic was obtained as described in the preceding section. However, in Section \ref{sec:theoretical_background}, a trade-off relationship between isotropy (quantization error minimization) and entropy maximization was revealed. Since entropy reduction degrades retrieval accuracy, a balance between isotropy and entropy should be kept. Accordingly, two methods for such balancing are proposed hereafter. The first one is the simpler one and  does not increase the number of fill-ins. The second one is using additional sparse rotation matrices. It increases the number of fill-ins, but has better accuracy than the first one in some cases.

\subsubsection{PCA tilting (PCAT)}  
In the first method, each pairwise rotation is "tilted" from the isotropic angle to the PCA angle (Fig. \ref{fig:ellipse}). It corresponds to $\theta = 0$ in equation (\ref{eq:r-theta_representation_of_2d_gauss}). Entropy is increased with this tilting, since the PCA angle is the entropy-maximizing angle. Rotation matrix is derived as
\begin{align} \label{eq:pca_rotation}
&R = 
\begin{pmatrix}
\cos \theta(\lambda) & -\sin \theta(\lambda) \\
\sin \theta(\lambda) & \cos \theta(\lambda)
\end{pmatrix}, \nonumber \\
\theta(\lambda) = \theta_{\text{iso}} + \lambda (&\theta_{\text{pca}} - \theta_{\text{iso}}), \quad \theta_{\text{pca}} = \tan^{-1} \left( \frac{1}{2} \frac{\sigma_{12}}{\sigma_{11} - \sigma_{22}}\right),
\end{align}
where $\theta_{\text{iso}}$ is given in Eq. (\ref{eq:iso_rotation}).

$\lambda$ is a tuning parameter ranging from zero (completely isotropic) to one (completely PCA). We can control a degree of balance between isotropy and entropy by tuning $\lambda$. Since PCA tilting does not lead to completely isotropic variances, there is no definite reason to stop applying basic rotations at $\log_2 n$ times. However, it is not necessary increase the number of basic rotations because it practically leads enough accuracy with $\log_2 n$ times application. The number of fill-ins of the transformation matrix therefore does not need to change. 

\subsubsection{Random Sparse PCA Rotation (RSPCA)}
The second method applies additional sparse rotations after having a completely isotropic transformation. The additional matrices have the same form as basic isotropic rotation, but $n/2$ rotational pairs are randomly chosen and PCA rotation ($\theta_{\text{pca}}$ given in Eq. (\ref{eq:pca_rotation})) is applied to each pair. This procedure is called "basic PCA rotation". 
Although it is not obvious how many times the basic PCA rotation should be applied, the experiments discussed in succeeding section show that $\mathrm{O}(\log_2 n)$ times rotations attains maximal retrieval accuracy. So the increasing of the number of fill-ins of the transformation matrix is very little.

\subsection{Relation with Major Existing Strategies}
The proposed algorithm introduces a novel strategy, in which the transformation matrix is expressed as a factored form of pairwise rotational matrices. For constructing each rotation, only the variance-covariance matrix is used. In contrast, some existing linear binary hashing algorithms (such as ITQ) use an objective function that is directly calculated from the data (e.g, quantization error due to discretization). These data-dependent objective functions capture non-gaussian property of the distribution of data. 

On the other hand, an arbitrary probability distribution function has an expansion series with the lowest order term given by a gaussian distribution. Such expansion is called Edgeworth expansion \cite{Kolassa2006}. From this viewpoint, it can be regarded as the lowest-order approximation is taken in our algorithm, whereas ITQ and other data-dependent methods consider higher order non-gaussian terms. Omission of higher order terms enables analytical treatments, which can provide a simple and computationally efficient binarization procedure. Despite the fact that the higher order terms are disregarded, the proposed method still achieves considerably high accuracy as explained below. 

\section{Experiments}\label{sec:experiments}
In the experiments, 128-dimensional gaussian toy data, 128-dimensional SIFT data, and high-dimensional VLAD data with various dimensions are used. The gaussian data is used for evaluating the theoretical behavior of the proposed algorithm. The SIFT data is used for comparing existing methods that is not feasible in high dimensions. The VLAD data is used for evaluating the algorithm in comparison with the state-of-the-art high-dimensional method.

\subsection{Experimental Protocols} \label{sec:experimental_protocols}
\subsubsection{Settings}
We use Top-10 recall as performance measure of binary hashing. Euclidean nearest neighbors in original feature space is used as ground truth. For the gaussian data, 10K data points for training, 2K for query and 100K for database is used. For SIFT data, we use SIFT1M dataset\cite{Jegou2011} and obey the original protocol (100K training set, 10K query set, and 1M database set). For creating VLAD data, we use ILSVRC2010 dataset \cite{Deng2009}. 25600-dimensional and 64000-dimensional VLAD is calculated from original SIFT data. 20K points for training and 5K points for queries are then randomly picked. The rest of the dataset (about 1M points) is used for the database.
 
\subsubsection{Existing methods to be compared}
We choose counterpart methods as follows: 
\textbf{Sparse Random Rotation (SRR)}: This is a random method corresponding to our sequence of sparse matrices scheme. The transformation matrix  for SRR has the same form as our method (Fig. \ref{illustration:basic_rotation}), whereas there is no sorting and rotation angle for each pair is randomly chosen. The number of basic rotation applied is set to $\lceil \log_2 n \rceil$.
\textbf{Iterative Quantization (ITQ)}\cite{Gong2013b}: This is one of the most well-known methods that keeps nearly-state-of-the-art performance for a wide range of data. It is considered as a reasonable performance counterpart for low-dimensional case.
\textbf{Isotropic Hashing (ISO)}\cite{Kong2012}: This is the original method that generates orthogonal transformation to make variances completely isotropic. For high-dimensional case ($d \ge 3$), there are generally an infinite number of isotropic states for any variance-covariance matrix. Each isotropic state has different retrieval performance because it differs from others in terms of entropy and higher-order cumulants (non-gaussian effects). It is considered as a counterpart that measures the quality of our isotropic transformation. We use Lift-and-Projection optimization algorithm proposed in \cite{Kong2012}.
\textbf{PCA Hashing (PCA)}: PCA hashing, as its name suggests, uses linear transformation to PCA basis. As discussed in section \ref{sec:ent_qerr}, PCA basis is the opposite extreme of the isotropic basis with regard to the trade-off relationship between quantization error and entropy. 
\textbf{K-means Hashing (KMH)}\cite{He2013}: This is a recently proposed state-of-the-art method. It uses k-means Vector Quantization and binary code assignment optimization for each cluster center. It is thus a kind of nonlinear method. It is selected for evaluating binary hashing performance compared to nonlinear methods. We use algorithm parameter $b=4, M=\text{ndim}/b$ and 50 iteration number defined in \cite{He2013}.
\textbf{Bilinear Projection-based Binary Codes (BPBC)}\cite{Gong2013}: This is the state-of-the-art high-dimensional hashing method using bilinear transformation. It is considered the baseline method. We use algorithm parameter $d_1 = 128, d_2 = \text{ndim} / d_1$, and 50 iteration number.

\subsection{Toy-data Experiment} \label{sec:toydata_experiment}

\begin{figure}[t]
\centering
\begin{tabular}{c}
\begin{minipage}{0.33\hsize}
	\centering
	\includegraphics[scale=0.14,angle=0]{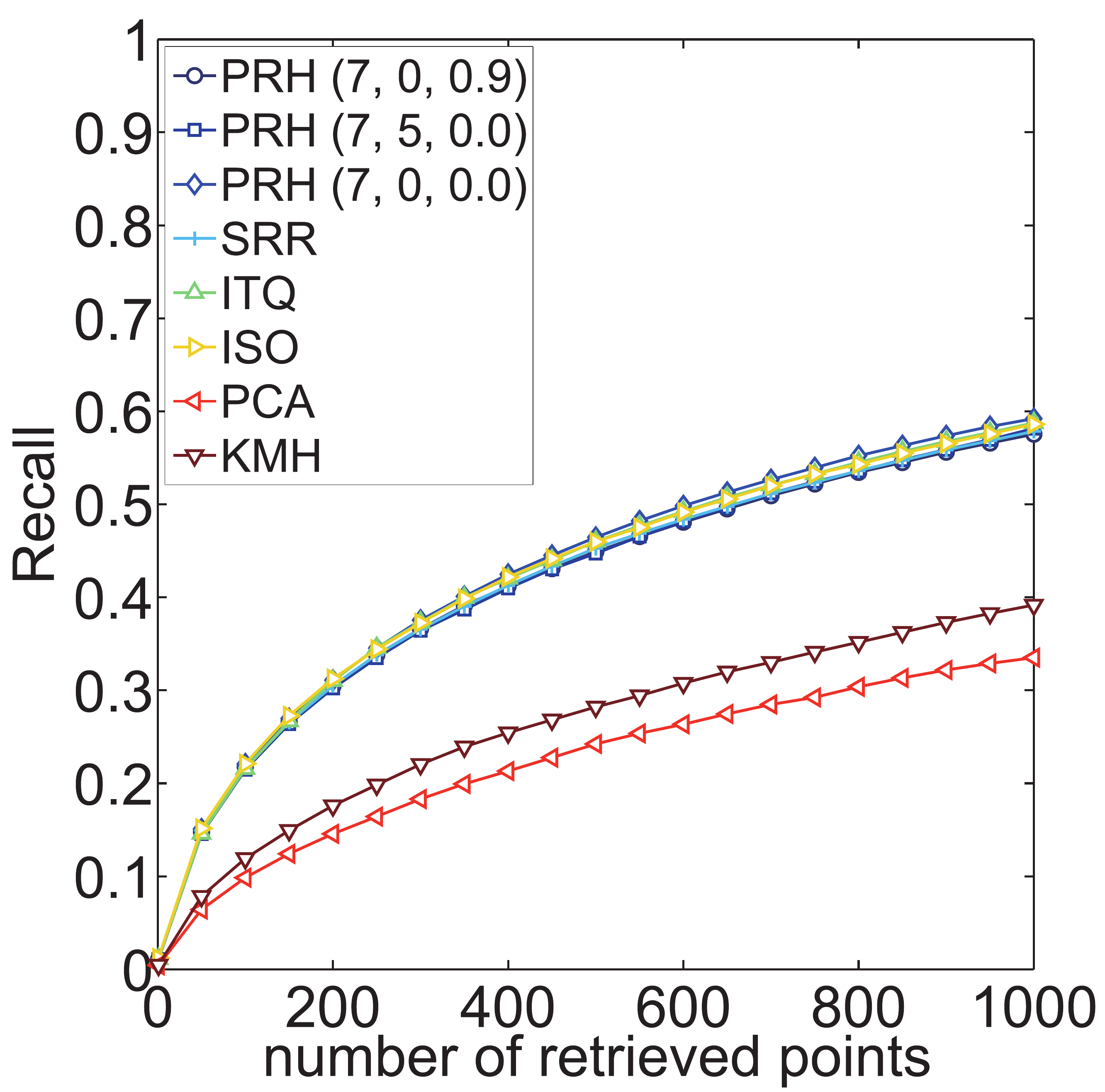}
	\hspace{0.5mm} (a) sphere-like gaussian. comparison with existing methods.
\end{minipage}
\begin{minipage}{0.33\hsize}
	\centering
	\includegraphics[scale=0.14,angle=0]{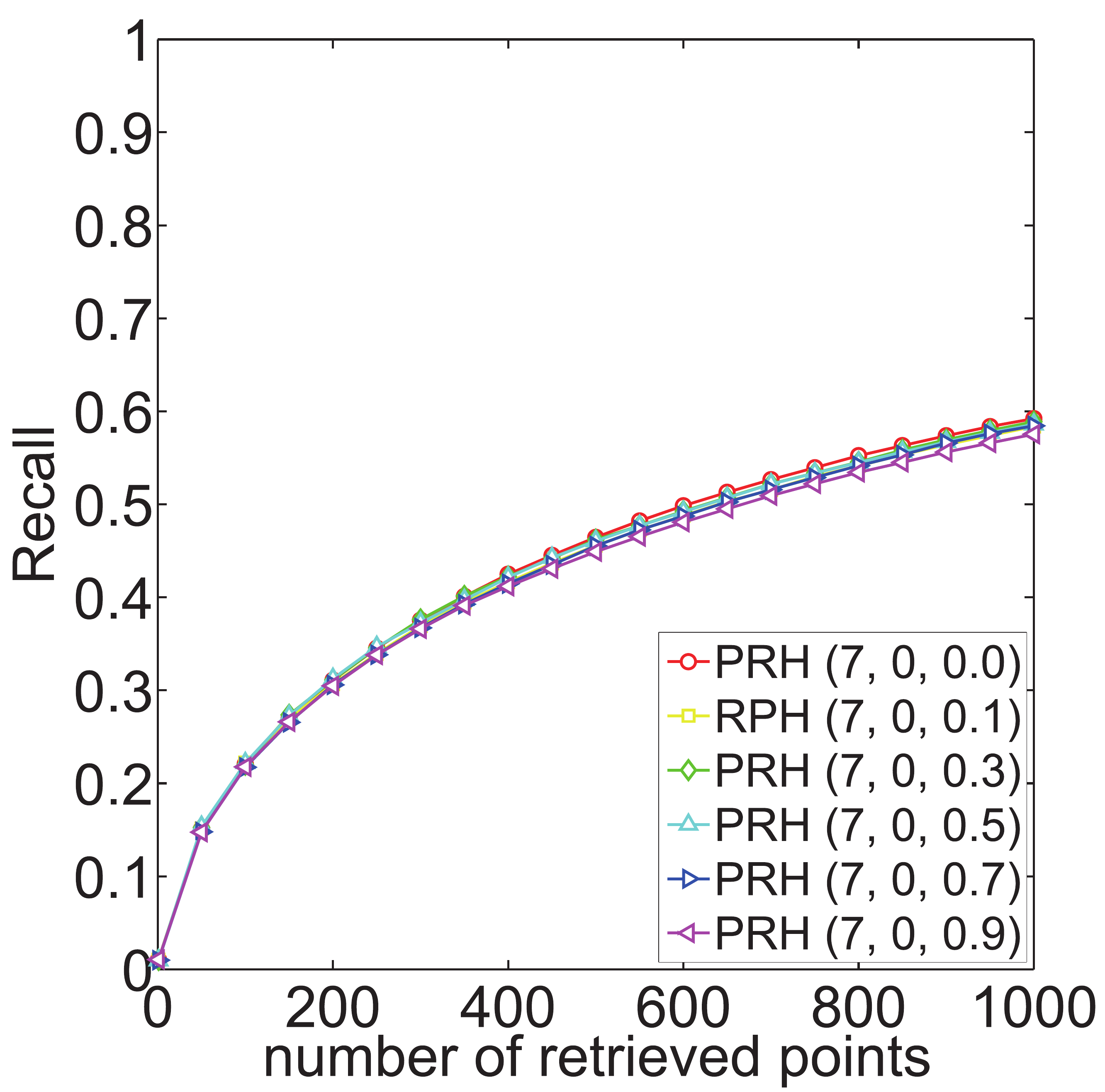}
	\hspace{0.5mm} (b) sphere-like gaussian. effect of PCAT. $\qquad \qquad \qquad$
\end{minipage}
\begin{minipage}{0.33\hsize}
	\centering
	\includegraphics[scale=0.14,angle=0]{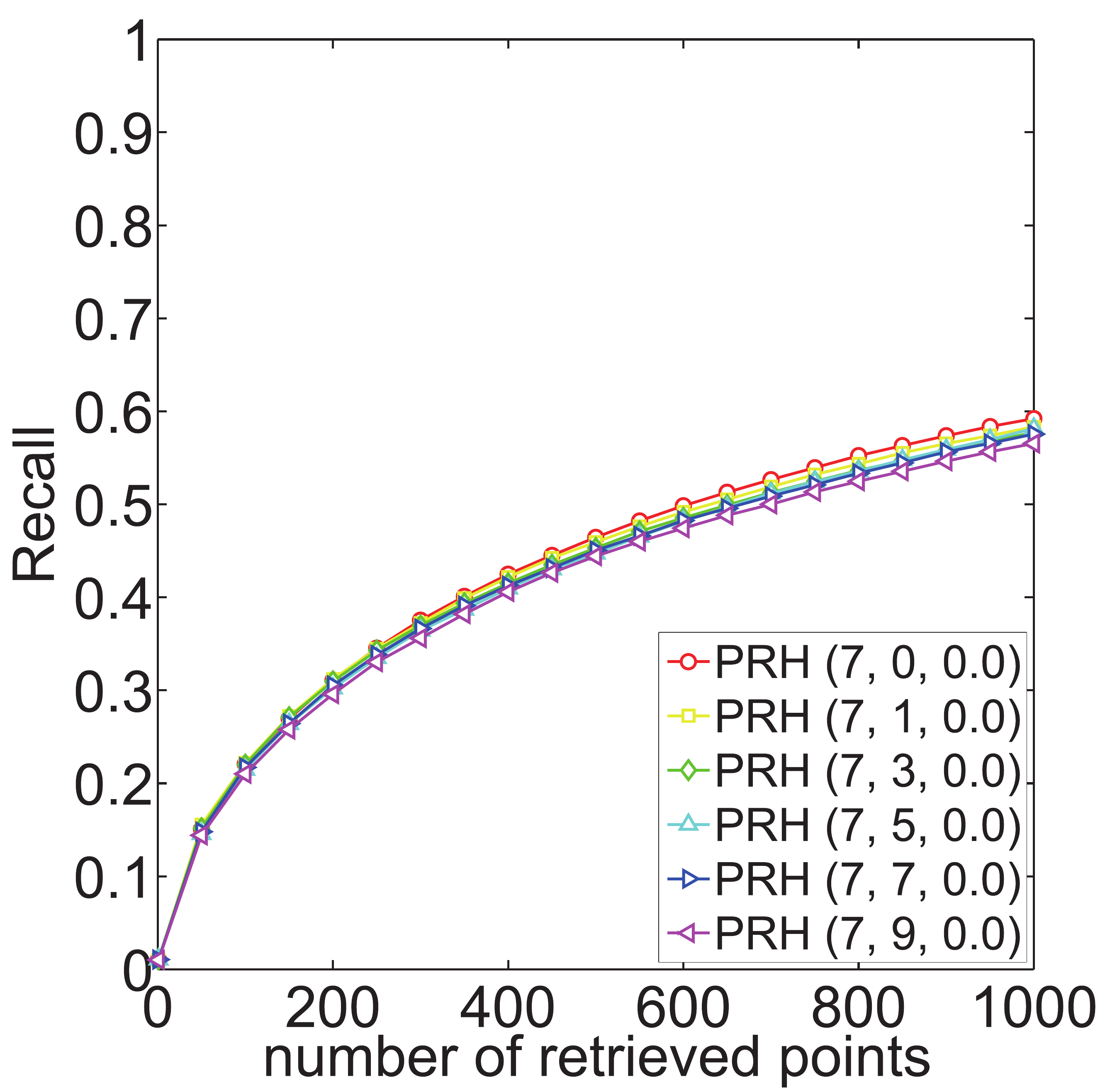}
	\hspace{0.5mm} (c) sphere-like gaussian. effect of RSPCA. $\qquad \qquad \qquad$
\end{minipage}
\end{tabular}
\begin{tabular}{c}
\begin{minipage}{0.33\hsize}
	\centering
	\includegraphics[scale=0.14,angle=0]{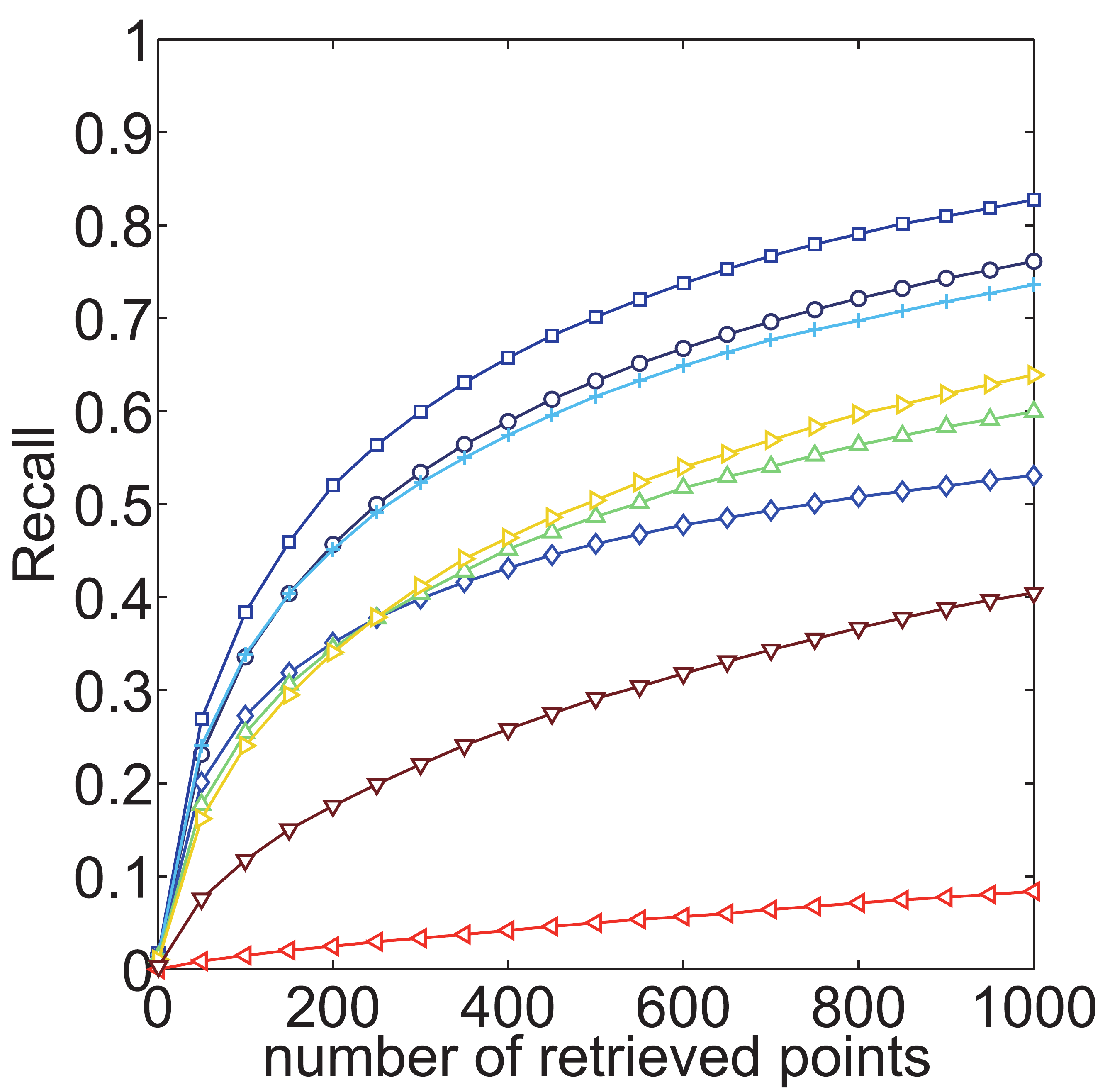}
	\hspace{0.5mm} (d) sharp gaussian. comparison with existing methods.
\end{minipage}
\begin{minipage}{0.33\hsize}
	\centering
	\includegraphics[scale=0.14,angle=0]{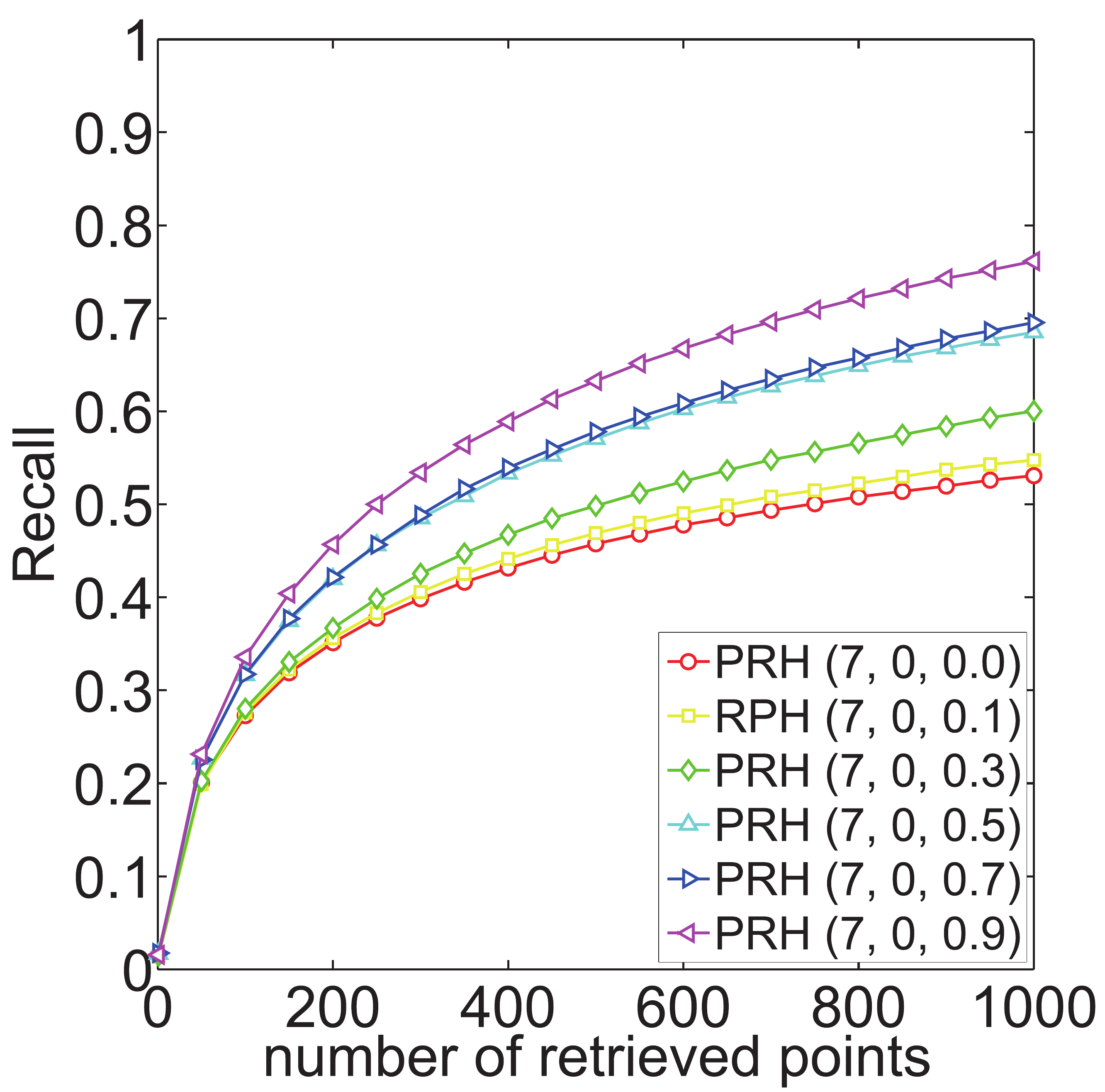}
	\hspace{0.5mm} (e) sharp gaussian. effect of PCAT. $\qquad \qquad \qquad \qquad \qquad$
\end{minipage}
\begin{minipage}{0.33\hsize}
	\centering
	\includegraphics[scale=0.14,angle=0]{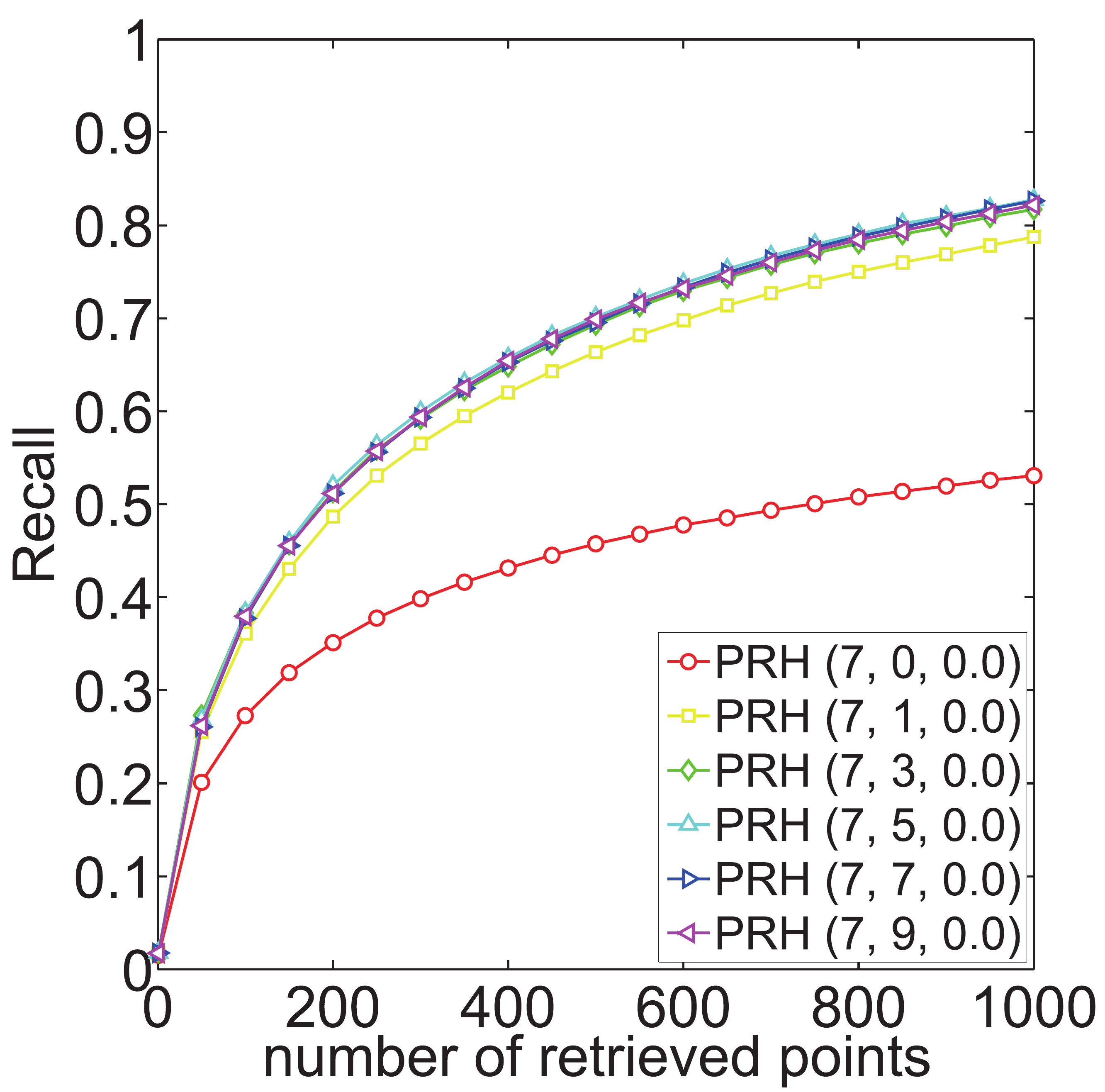}
	\hspace{0.5mm} (f) sharp gaussian. effect of RSPCA. $\qquad \qquad \qquad \qquad \qquad$
\end{minipage}
\end{tabular}
\caption{Top-10 NN retrieval results for 128-dimensional gaussian data. PRH(m, n, $\lambda$) indicates the proposed method with $m$-times basic isotropic rotation, $n$-times basic PCA rotation, and PCAT parameter $\lambda$ (Eq. (\ref{eq:pca_rotation})). Upper row is for the data with log-variance value of one. Lower row is for the data with log-variance of three (described in Section \ref{sec:toydata_experiment}). Abbreviated legends of plot (d) are the same as that of plot (a). \label{experiment:gauss}}
\end{figure}

First we use artificial gaussian data to observe theoretical behavior of the proposed algortihm, which is discussed above.

A 128-dimensional random variance-covariance matrix is created and used to generate mean-centered gaussian data. To create a variance-covariance matrix, a diagonal matrix with random positive eigenvalues that is distributed log-normally is generated. Then a diagonal matrix is rotated by random rotation. We consider two different eigenvalue distributions. One uses a log-normal distribution with log variance of one (sphere-like distribution), and the other uses a log-normal distribution with log variance of three (sharp distribution).

Fig. \ref{experiment:gauss} shows the retrieval results. In the case of a sphere-like distribution (upper row), most of methods have little difference in accuracy because the shape of the distribution is insusceptible under rotational transformation. A notable point is that in the case of sharp distribution (lower row), completely isotropic PRH is obviously inferior to the other cases, although Isotropic Hashing, which also has completely isotropic variances, achieves reasonable performance. As discussed in Section \ref{sec:experimental_protocols}, there are an infinite number of isotropic states. The Lift-and-Projection in Isotropic Hashing tends to find entropically favorable isotropic states. Despite the fact that PRH is extremely simple and sparse, it sometimes achieves entropically inferior isotropic states. However, this inferiority is reasonably overcome by PCAT or RSPCA without loss of sparsity.

The lower middle plot of Fig. \ref{experiment:gauss} indicates that almost-PCA angle ($\lambda \sim 1$) at each pairwise rotation leads to good performance in the sharp gaussian distribution. It is important to distinguish our sequential pairwise almost-PCA rotation and PCA hashing rotation. To obtain exact PCA basis, it is necessary to account for all $n(n-1)/2$ possible pairs. PCAT, however, only deals with $\mathrm{O}(n \log_2 n)$ pairs.

\subsection{Real Datasets}
\subsubsection{Low Dimensional Case}

\begin{figure}[t]
\centering
\begin{tabular}{c}
\begin{minipage}{0.33\hsize}
	\centering
	\includegraphics[scale=0.14,angle=0]{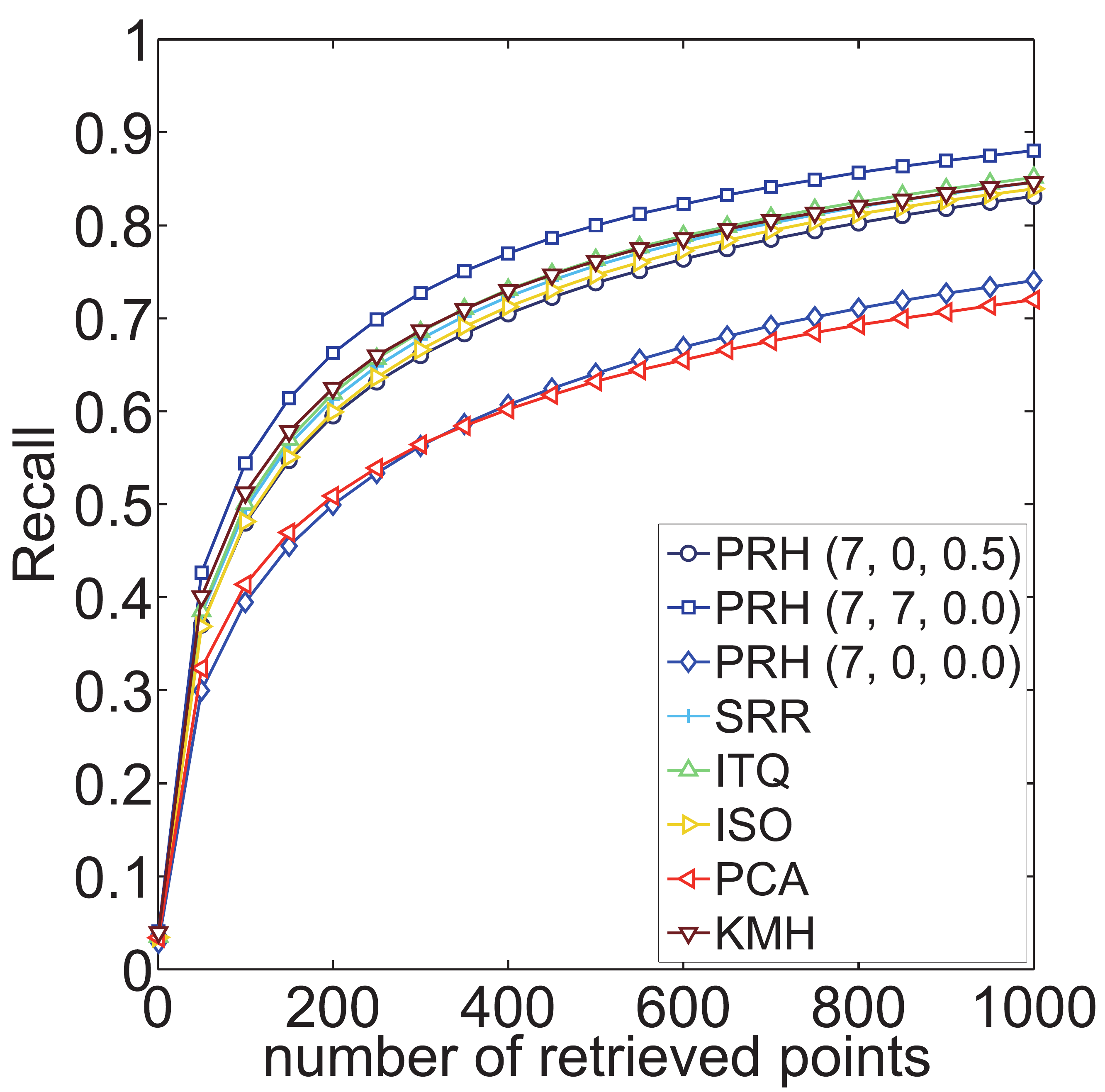}
	\hspace{0.5mm} (a) comparison with existing methods.
\end{minipage}
\begin{minipage}{0.33\hsize}
	\centering
	\includegraphics[scale=0.14,angle=0]{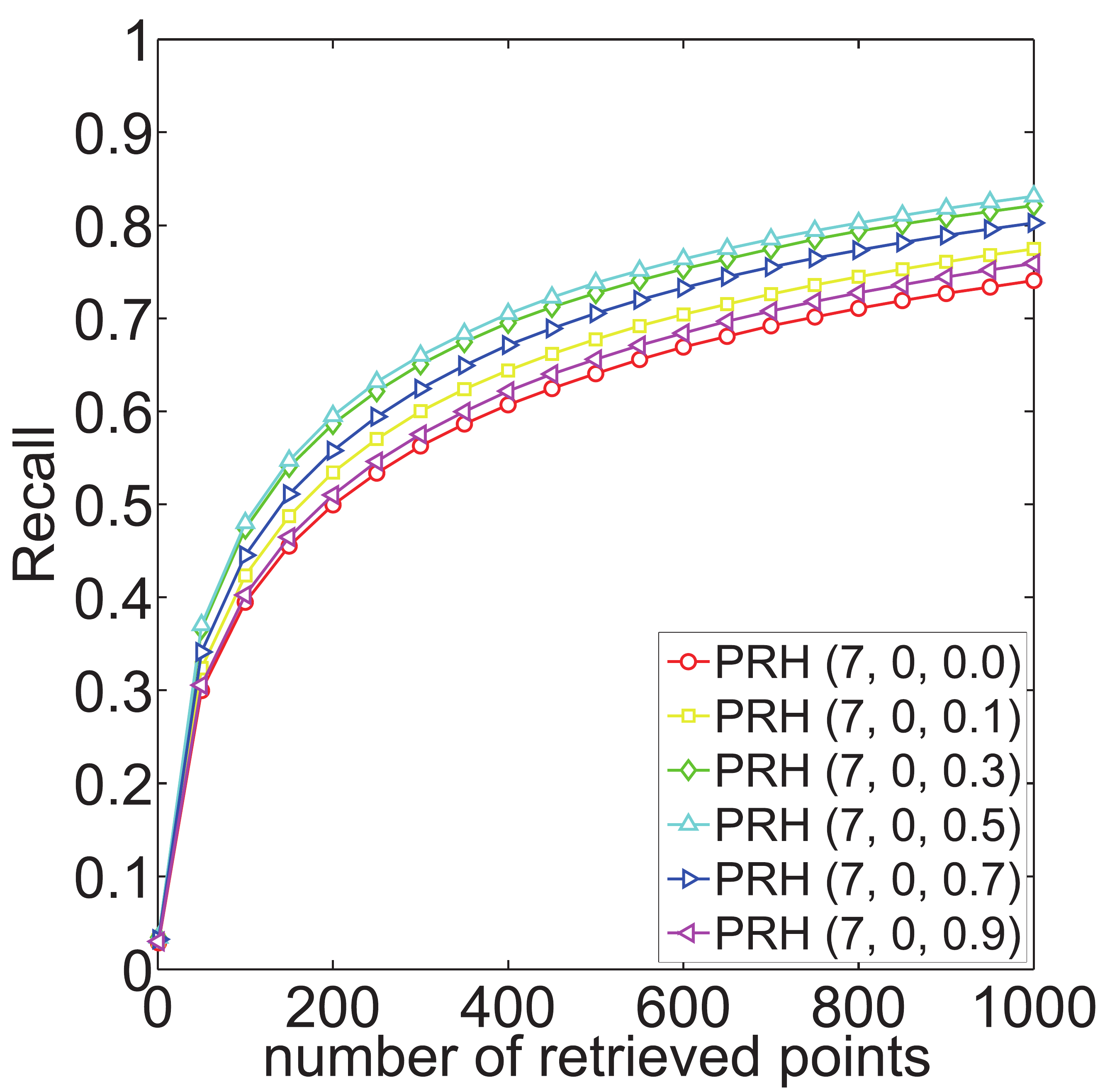}
	\hspace{0.5mm} (b) effect of PCAT. $\qquad \qquad \qquad$
\end{minipage}
\begin{minipage}{0.33\hsize}
	\centering
	\includegraphics[scale=0.14,angle=0]{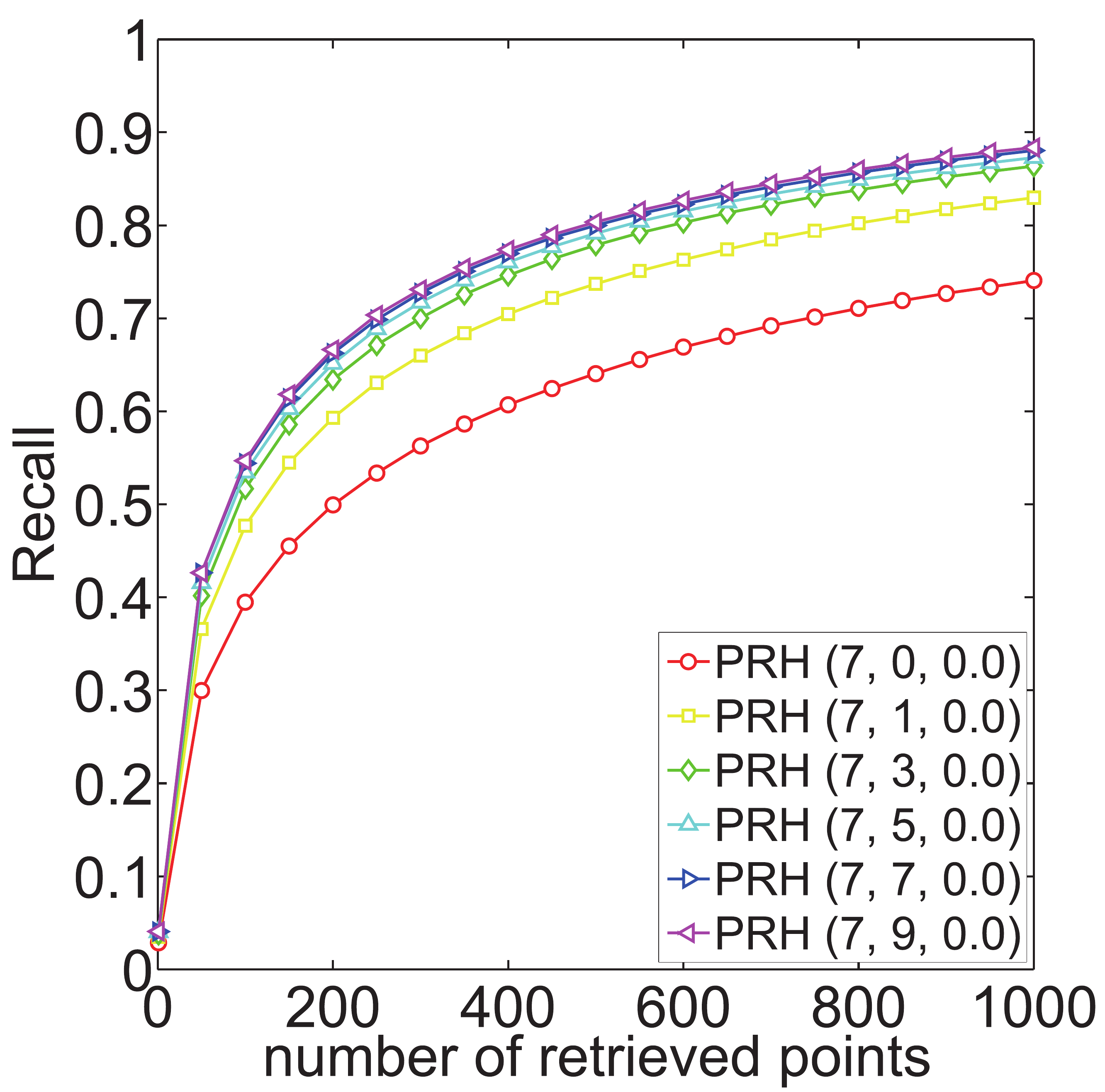}
	\hspace{0.5mm} (c) effect of RSPCA. $\qquad \qquad \qquad$
\end{minipage}
\end{tabular}
\caption{Top-10 NN retrieval results of 128-dimensional SIFT1M data. The meaning of PRH(m, n, $\lambda$) is shown in Figure \ref{experiment:gauss}.}
\label{experiment:sift1m}
\end{figure}

\begin{figure}[t]
\centering
\begin{tabular}{c}
\begin{minipage}{0.33\hsize}
	\centering
	\includegraphics[scale=0.14,angle=0]{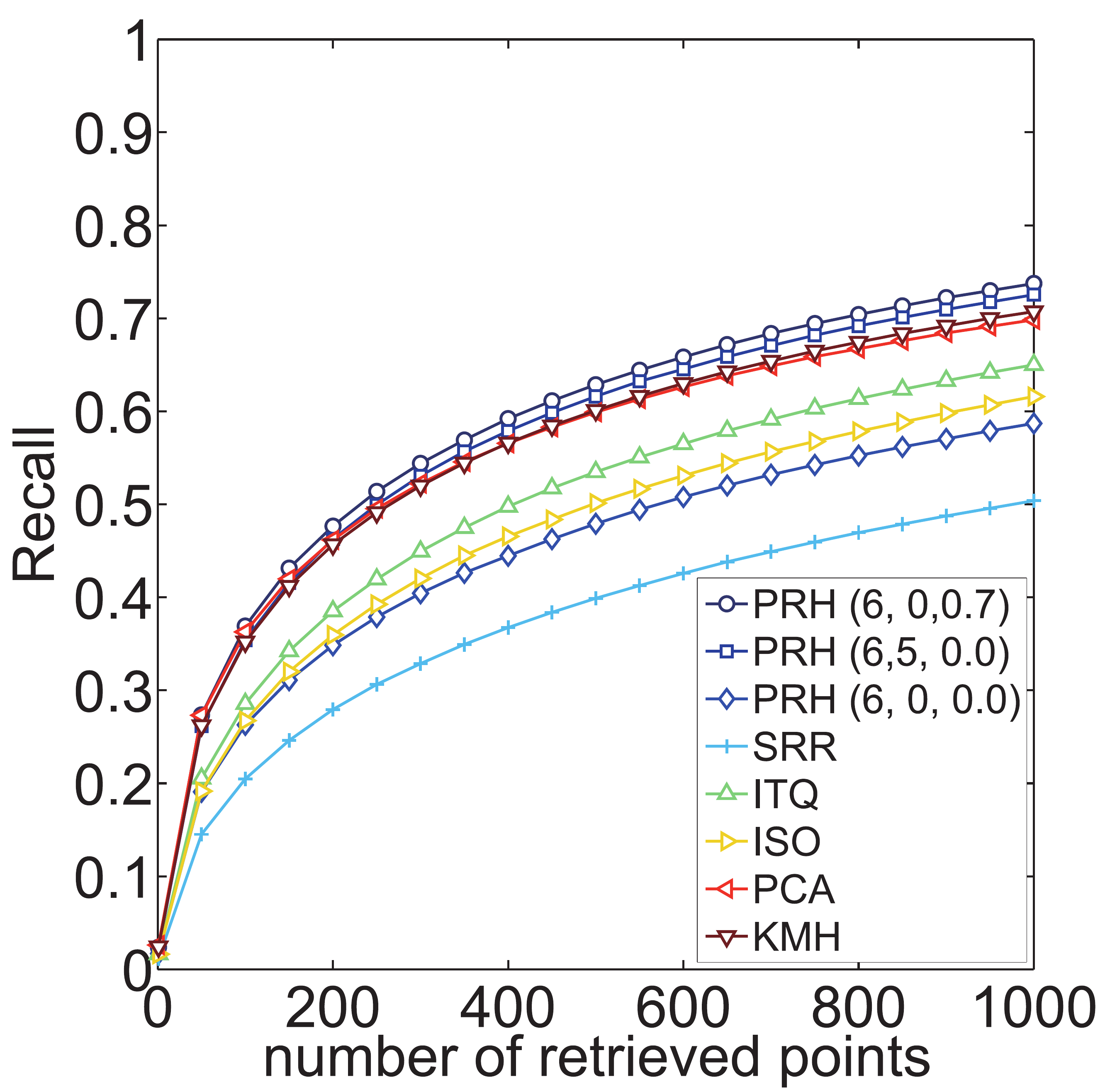}
	\hspace{0.5mm} (a) 64bit (1/2 dimension reduction) results.
\end{minipage}
\begin{minipage}{0.33\hsize}
	\centering
	\includegraphics[scale=0.14,angle=0]{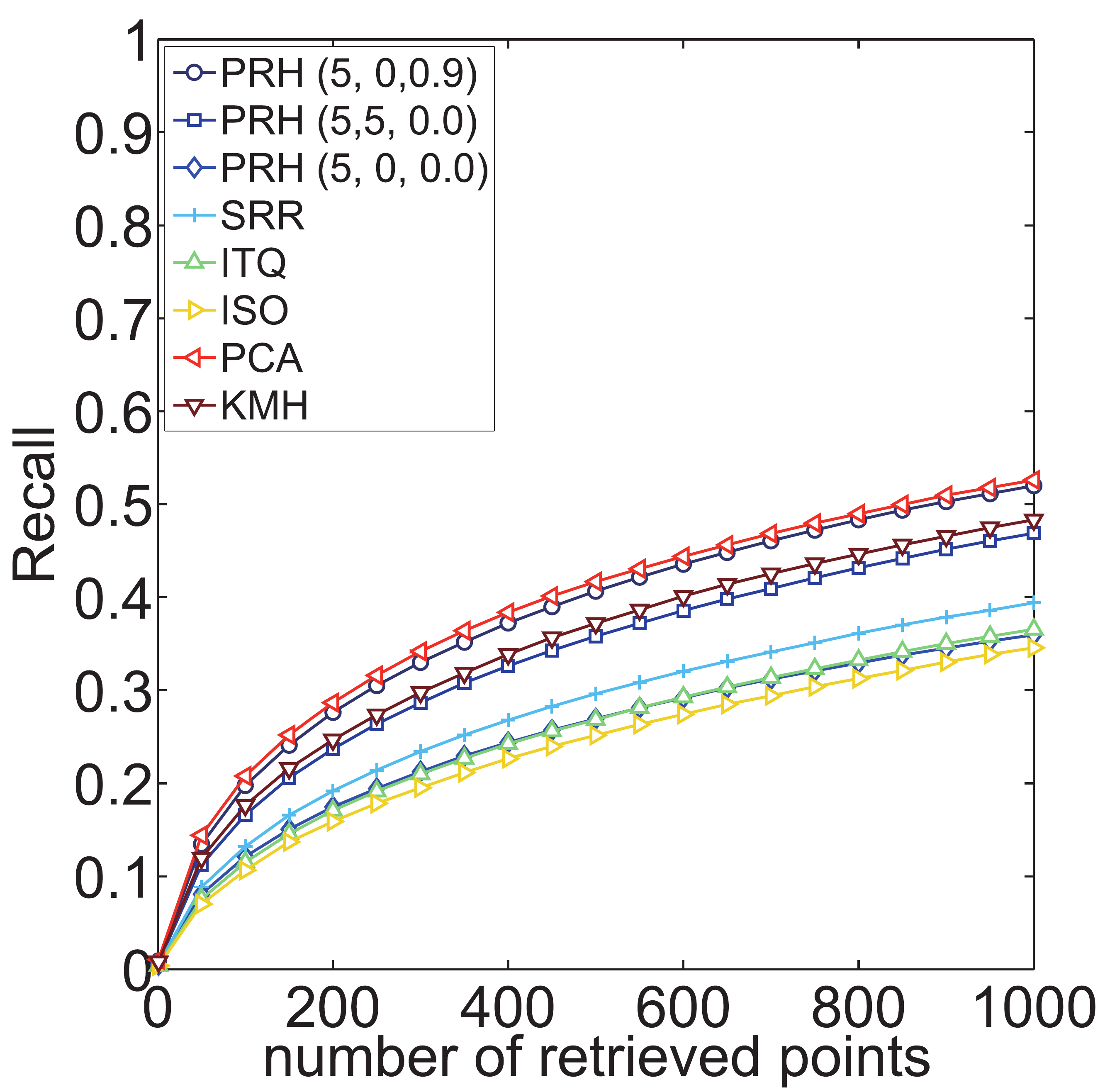}
	\hspace{0.5mm} (b) 32bit (1/4 dimension reduction) results.
\end{minipage}
\begin{minipage}{0.33\hsize}
	\centering
	\includegraphics[scale=0.14,angle=0]{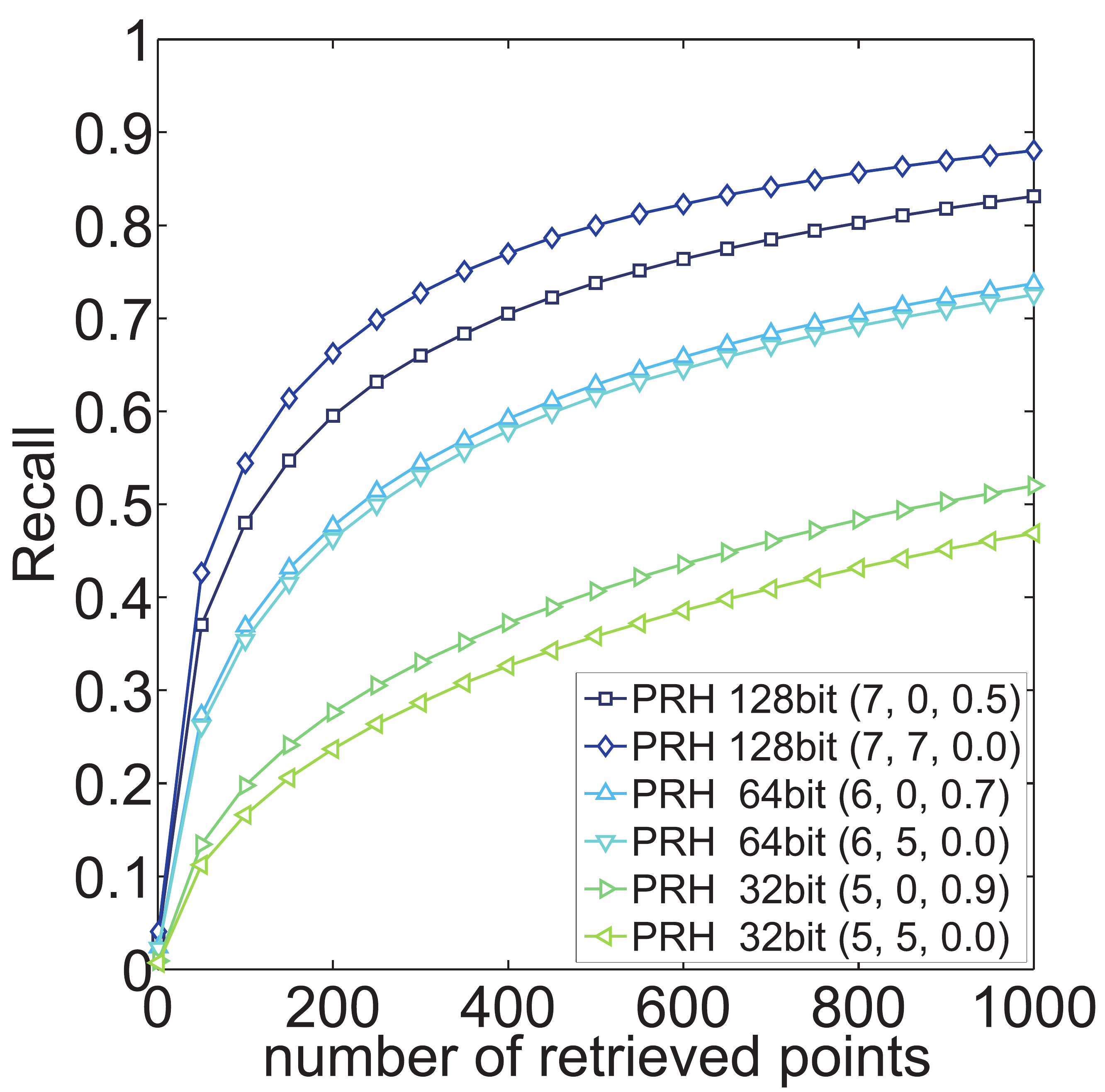}
	\hspace{0.5mm} (c) PRHs for different degree of reductions.
\end{minipage}
\end{tabular}
\caption{Top-10 NN retrieval results of SIFT1M data with PCA dimension reduction. \textbf{Left}:64bit (1/2 reduction) case. \textbf{Middle}: 32bit (1/4 reduction) case. \textbf{Right}: behavior of PRHs across some dimensions. The meaning of PRH(m, n, $\lambda$) is shown in Figure \ref{experiment:gauss}. \label{experiment:sift1m_reduct}}
\end{figure}  

The SIFT1M case is considered next.
Fig. \ref{experiment:sift1m} is the retrieval results. In common with the gaussian case, the completely isotropic PRH (PRH(7, 0, 0.0)) leads unfavorable accuracy. PCAT and RSPCA attain good performance. Especially, RSPCA achieves remarkably better retrieval result compared to other methods in figure (d). 

\subsubsection{Relation to dimension reduction}
Although a sparse dimension reduction scheme is not devised in this study, we examine the effect of dimension reduction on the performance of the proposed algorithm. As with existing methods, PCA basis is tentatively used as dimension reduction for PRH and SRR. Note that it does not keep sparsity of transformation. Fig \ref{experiment:sift1m_reduct} shows the results of SIFT1M. It is clear that the proposed algorithm maintains higher performance with each dimension reduction compared to the other methods.

\subsubsection{High-dimensional Case}

\begin{figure}[t]
\centering
\begin{tabular}{c}
\begin{minipage}{0.33\hsize}
	\centering
	\includegraphics[scale=0.14,angle=0]{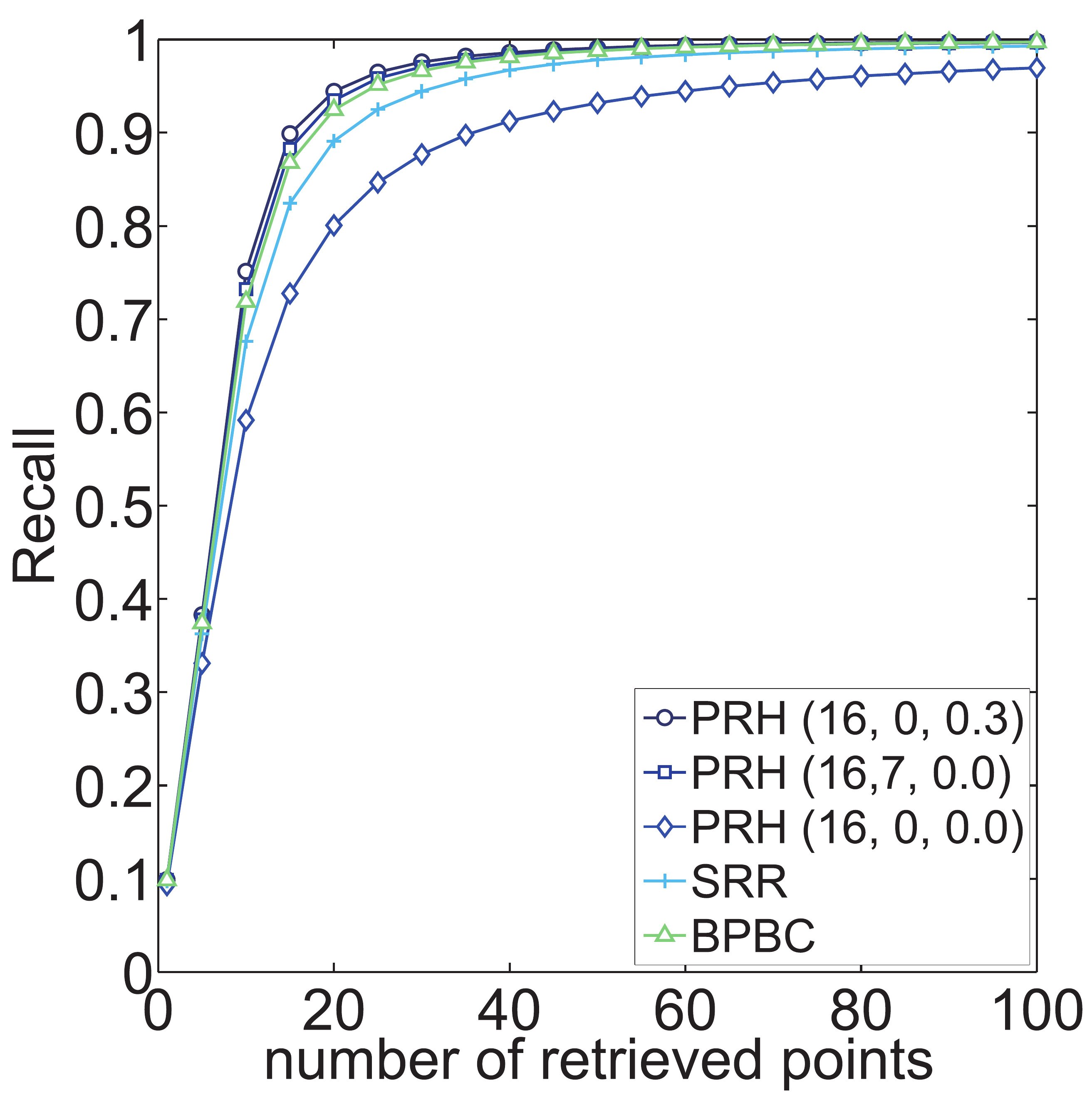}
	\hspace{0.5mm} (a) comparison with baseline method.
\end{minipage}
\begin{minipage}{0.33\hsize}
	\centering
	\includegraphics[scale=0.14,angle=0]{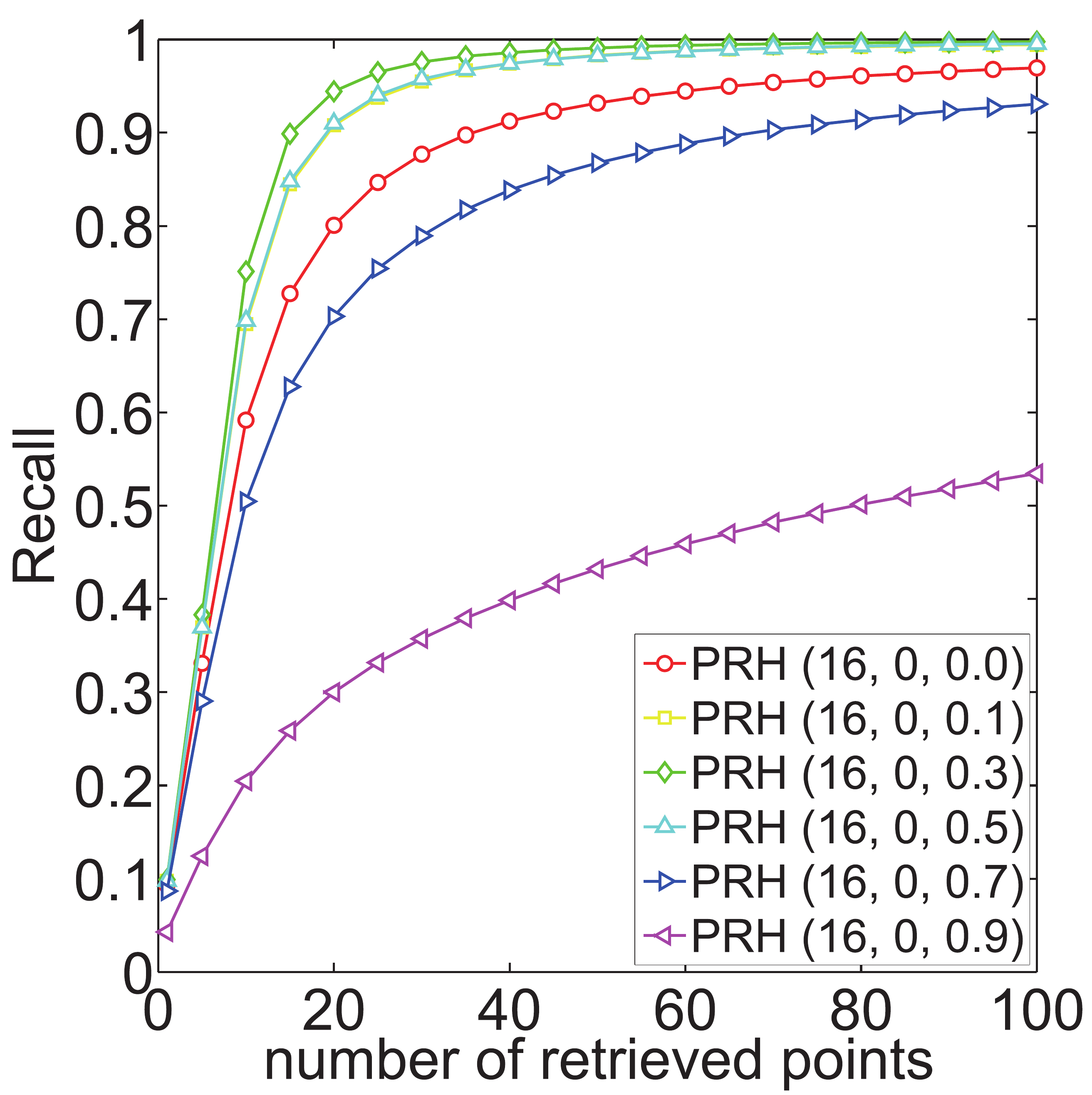}
	\hspace{0.5mm} (b) effect of PCAT. $\qquad \qquad \qquad$
\end{minipage}
\begin{minipage}{0.33\hsize}
	\centering
	\includegraphics[scale=0.14,angle=0]{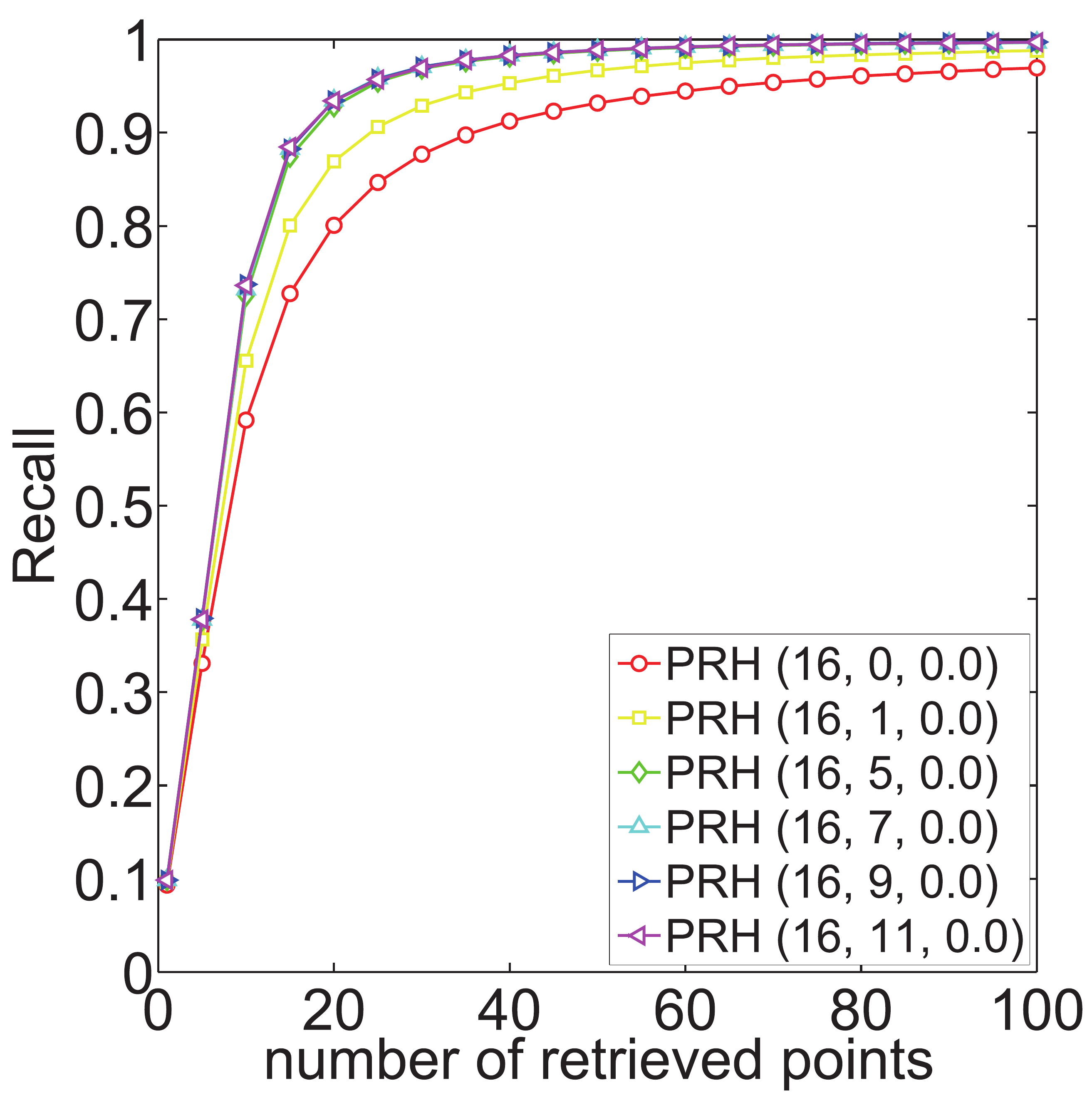}
	\hspace{0.5mm} (c) effect of RSPCA. $\qquad \qquad \qquad$
\end{minipage}
\end{tabular}
\begin{tabular}{c}
\begin{minipage}{0.33\hsize}
	\centering
	\includegraphics[scale=0.14,angle=0]{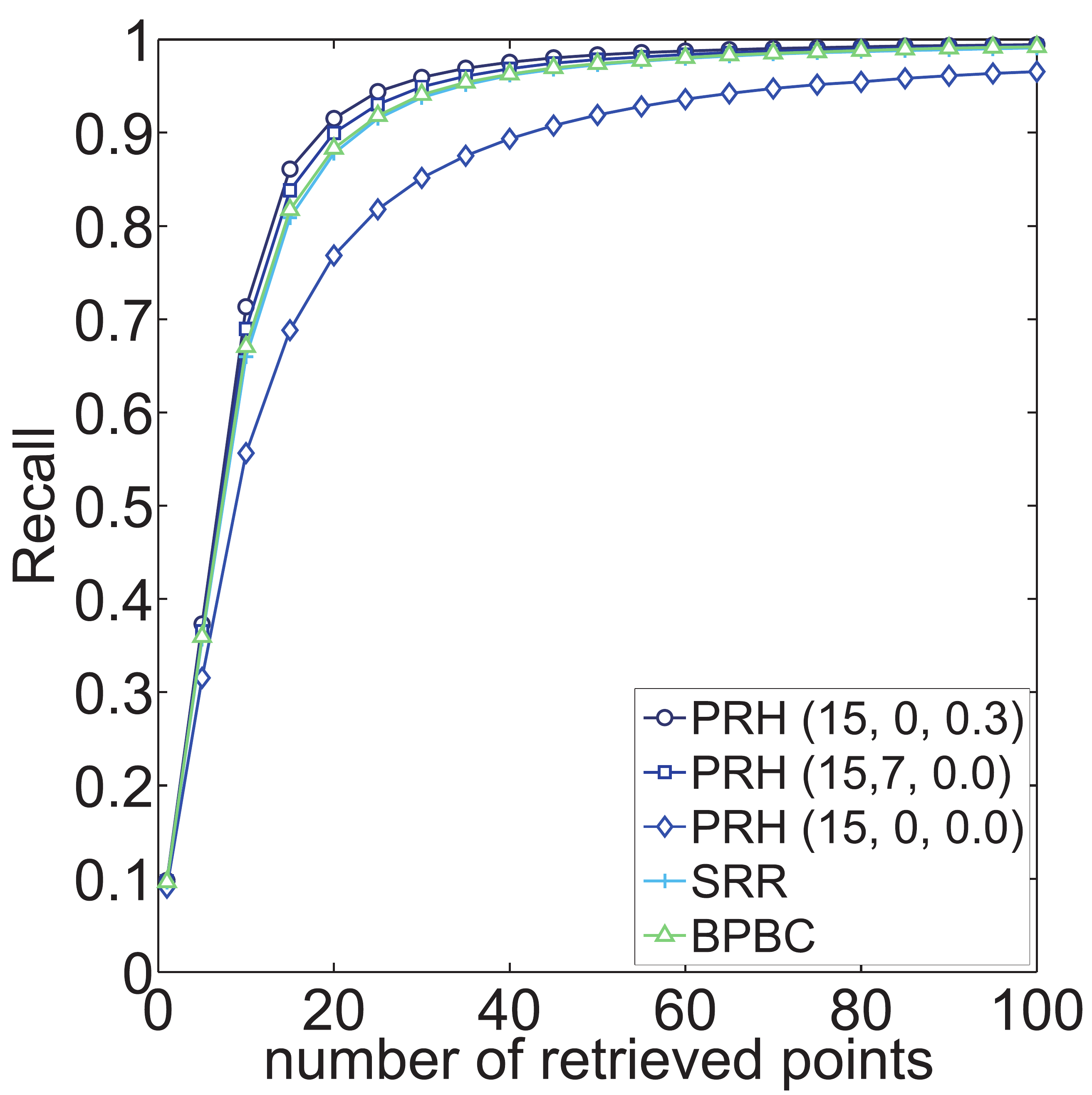}
	\hspace{0.5mm} (d) comparison with baseline method.
\end{minipage}
\begin{minipage}{0.33\hsize}
	\centering
	\includegraphics[scale=0.14,angle=0]{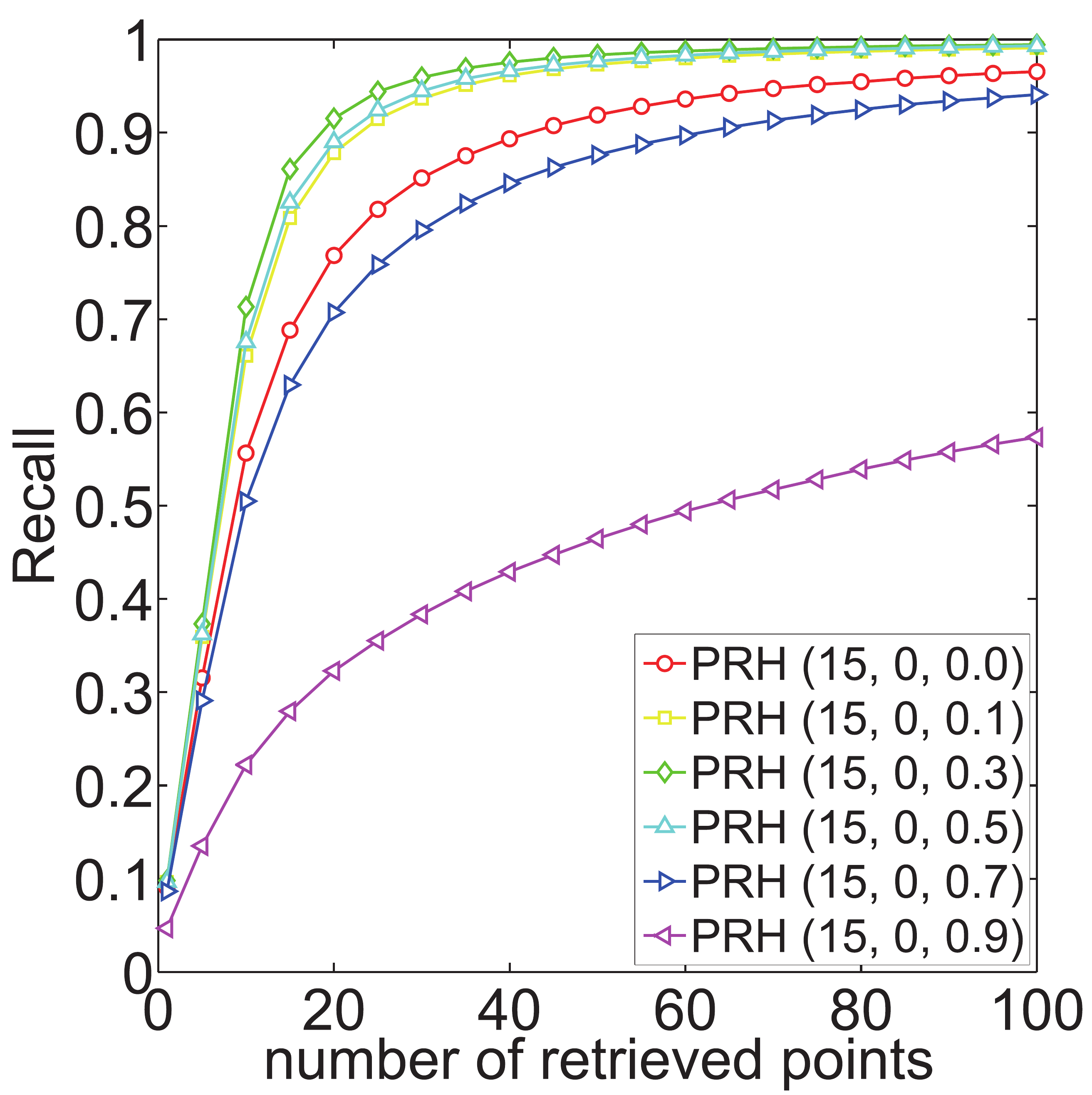}
	\hspace{0.5mm} (e) effect of PCAT. $\qquad \qquad \qquad \qquad \qquad$
\end{minipage}
\begin{minipage}{0.33\hsize}
	\centering
	\includegraphics[scale=0.14,angle=0]{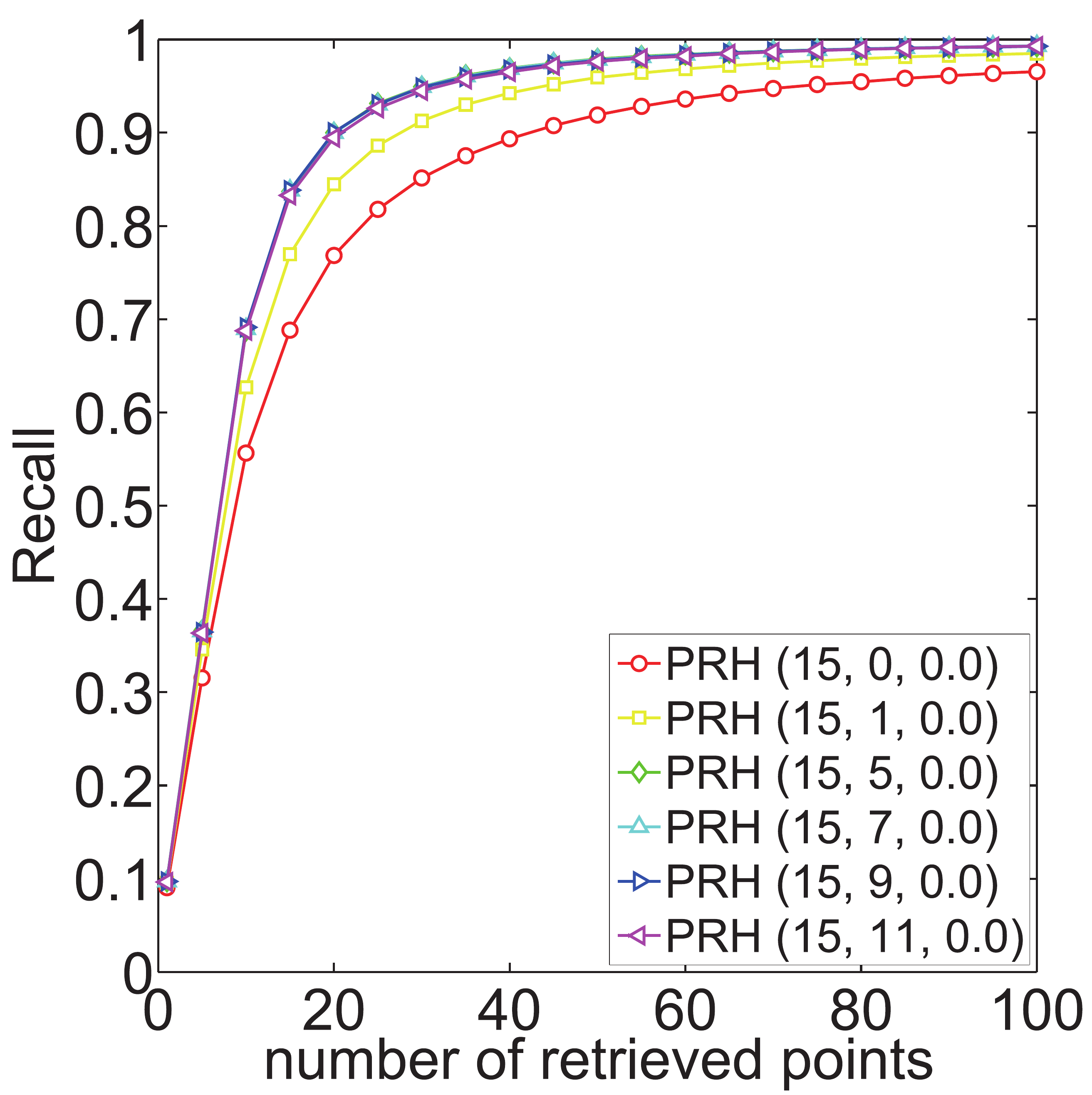}
	\hspace{0.5mm} (f) effect of RSPCA. $\qquad \qquad \qquad$
\end{minipage}
\end{tabular}
\caption{Top-10 NN retrieval results for 64000 (\textbf{Upper}) and 25600-dimensional (\textbf{Lower}) VLAD data. The meaning of PRH(m, n, $\lambda$) is shown in Figure \ref{experiment:gauss}. \label{fig:vlad64000}}
\end{figure}

The high-dimensional case, which is the main contribution of our algorithm, is examined next. Fig. \ref{fig:vlad64000} is the retrieval results for 64000-dimensional and 25600-dimensional VLAD features calculated from ILSVRC2010 dataset. PCAT achieves state-of-the-art retrieval accuracy. It is verified that it attains high performance for each dimension. In the experiment, contrary to the lower-dimensional case, RSPCA is inferior to PCAT. We think that this is the result of improper random pairing  in RSPCA  because the possible number of pairing is $O(n^2)$.

\subsection{Computational Cost}

\begin{table}[t]
 \centering
 \caption{Learning time for each methods. \textbf{Left}: Learning time for SIFT1M dataset. \textbf{Right}: Learning time for 25600-dimensional VLAD case. \label{table:learntime}}
 \scalebox{0.8}[0.8]{
  \begin{tabular}{|l||c|r|} \hline
    Methods & Learn t(s) \\ \hline \hline
    PRH(7,0,0.5) & 0.11\\ \hline
    PRH(7,7,0.0) & 0.10\\ \hline
    SRR & 0.015\\ \hline
    ITQ & 12.0 \\ \hline
    ISO & 3.90\\ \hline
    PCA & 0.09\\ \hline
    KMH & 402\\ \hline
  \end{tabular}
  }
  \scalebox{0.8}[0.8]{
  \hspace{3mm}
  \begin{tabular}{|l||c|r|} \hline
    Methods & Learn t(s) \\ \hline \hline
    PRH(15,0,0.3) & 344 \\ \hline
    PRH(15,15,0.0) & 527 \\ \hline
    SRR & 24\\ \hline
    BPBC & 1740\\ \hline
  \end{tabular}
  }
\end{table}

\begin{figure}[t]
\centering
\includegraphics[scale=0.14,angle=0]{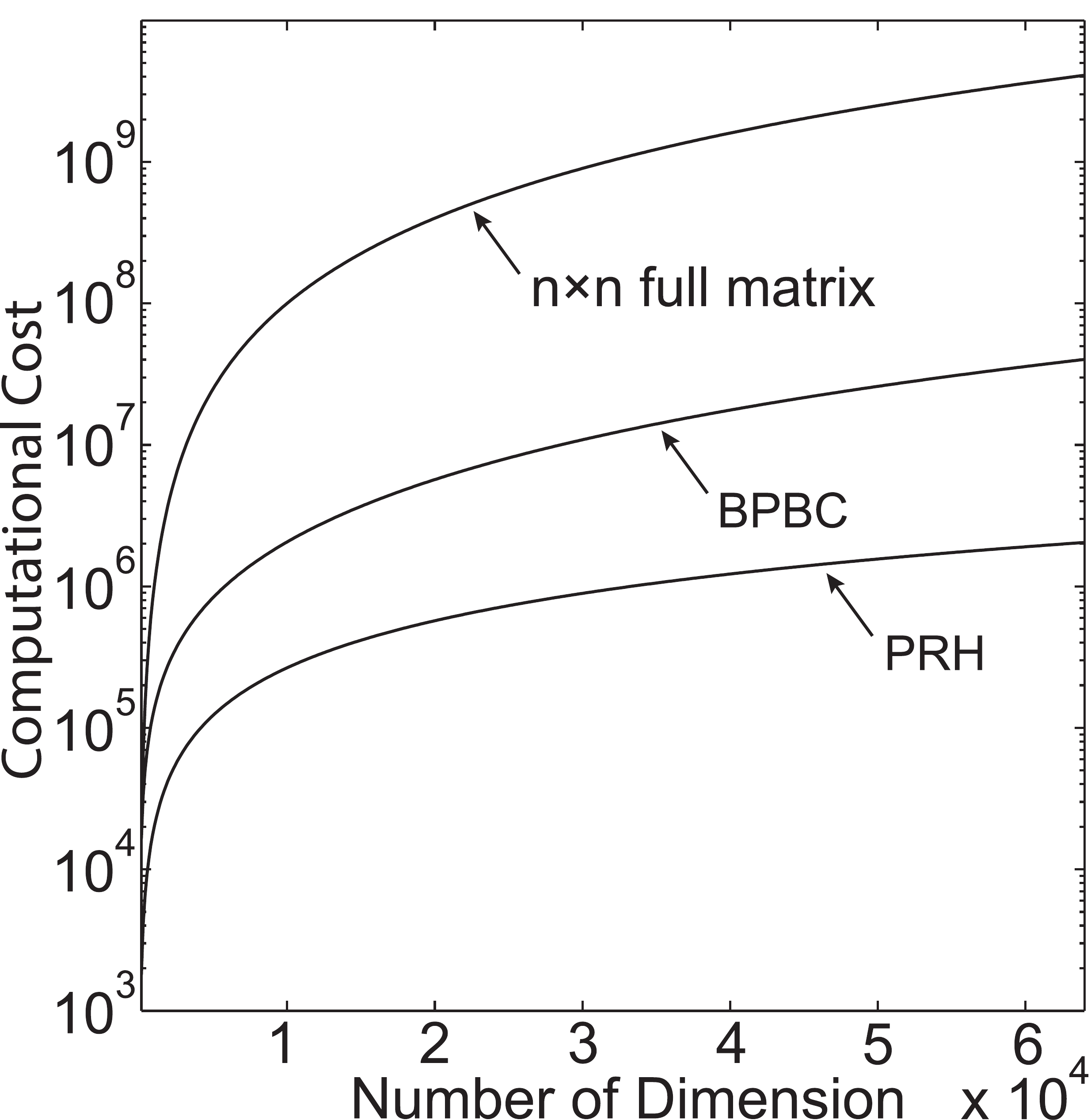}
\hspace{5mm}
\raisebox{-1mm}{\includegraphics[scale=0.14,angle=0]{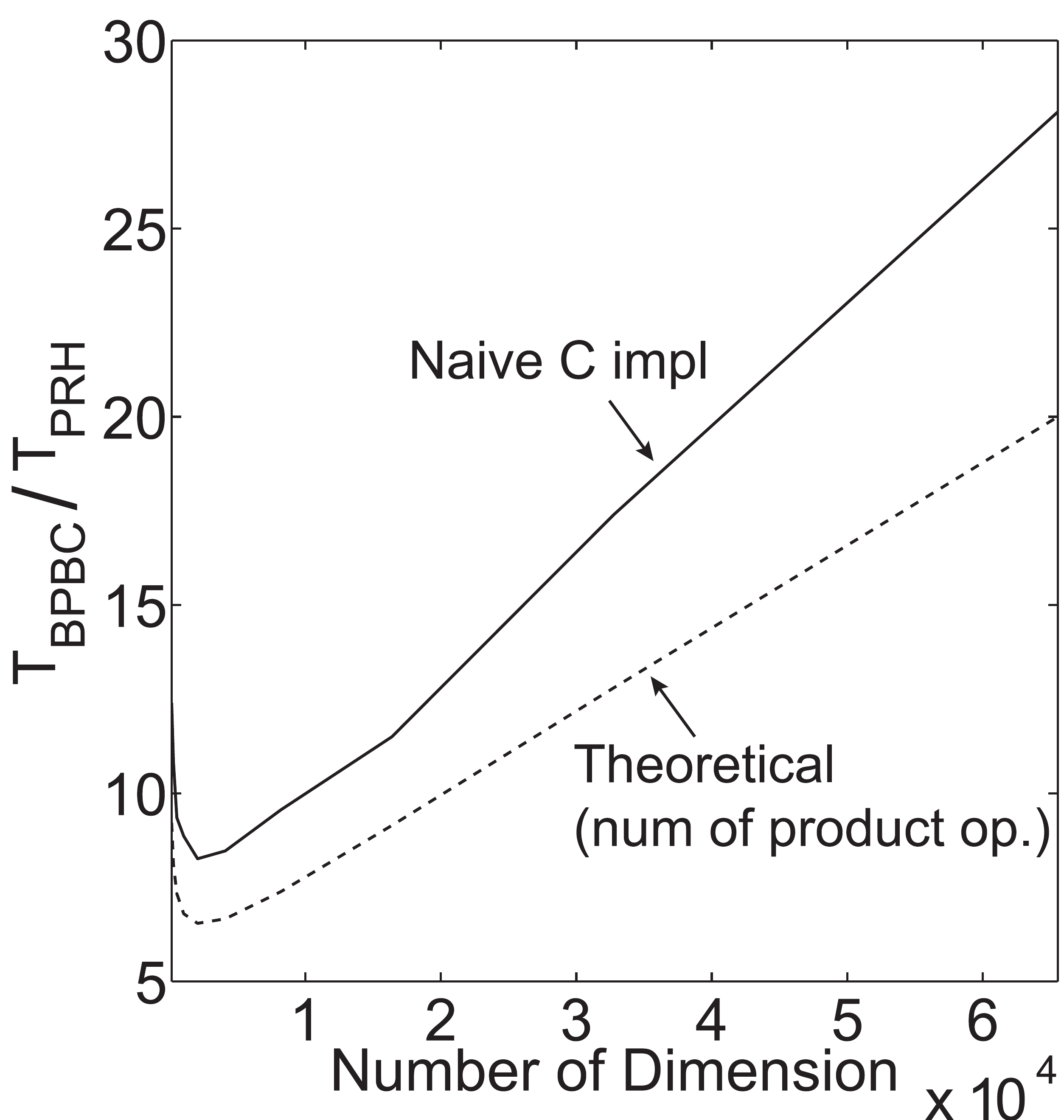}}
\caption{Encoding cost. \textbf{Left}: Comparison of the number of product operations. \textbf{Right}: Encoding speed improvement ratio of PRH in comparison with BPBC.  \label{fig:computationalcost}}
\end{figure}

Since our implementation is using a MATLAB sparse matrix datatype, it is difficult to reasonably evaluate encoding cost in comparison with other methods that use optimized dense matrix operations. We use tentative environment for the evaluation. We show the comparison of the number of product operations and speed improvement ratio to BPBC in encoding phase (Fig. \ref{fig:computationalcost}). The improvement ratio is calculated with naive C implementation of dense/sparse matrix operation and  it is compared with the theoretical (the number of product operations) one. The number of sum operation is also cut down in our method and we only need one sum operation per two product operations. BPBC needs almost the same number of sum and product operations. This attributes would be an explanation of the exceeding of naive C implementation result to theoretical speed improvement ratio.

Table. \ref{table:learntime} is the learning time comparison. PRH learns very fast in each case. For high dimensional case, 25600-dimensional learning time is shown to store all of the data on memory. Our implementation is not optimized (efficient treatment of sparse and symmetric matrix is possible).

\section{Conclusion}
We have proposed Pairwise Rotation Hashing (PRH), a linear binary hashing algorithm that has $\mathrm{O}(n \log n)$ encoding cost. PRH is based on two-dimensional analytical study of trade-off relationship between quantization error and entropy. The proposed algorithm is also fast in the learning phase because it needs only $\mathrm{O}(n \log n)$ computations in the iteration loop. It shows high hashing accuracy in retrieval tasks at both low and high dimensions. Especially it achieves state-of-the-art performance at high dimensions (10K or higher).

We still have room for improvement. In this study, a dimension reduction scheme compatible with pairwise concept is excluded. We have an idea that dimension reduction can be done again in a pairwise fashion, i.e, droping minor components in the pairwise PCA in Eq. \ref{eq:pca_rotation}. A key issue is to find an appropriate pairing method in the pairwise PCA part.
Even though RSPCA demonstrated high performance, it can be further improved if components are paired in selective ways, rather than in random ways. In RSPCA, there is the potential for finding more sophisticated pairing scheme that favorably balances isotropy with entropy. However, the exhaustive search of $\mathrm{O}(n^2)$ possible pairing substantially degrades the learning speed. Non-random but efficient pairing scheme is needed.

\bibliographystyle{splncs}
\bibliography{test2}{}
\end{document}